\documentclass{article}

\usepackage[preprint]{neurips_2025}

\usepackage[utf8]{inputenc} 
\usepackage[T1]{fontenc}    
\usepackage{amsmath,amsthm,amssymb} 
\usepackage{hyperref}       
\usepackage{cleveref}      
\usepackage{url}            
\usepackage{booktabs}       
\usepackage{amsfonts}       
\usepackage{nicefrac}       
\usepackage{microtype}      
\usepackage[dvipsnames]{xcolor}         
\usepackage{graphicx}
\usepackage{subfigure}
\usepackage{booktabs} 
\usepackage{bm}
\usepackage{physics2}
\usephysicsmodule{ab, ab.braket, doubleprod, diagmat, xmat}
\usepackage{fixdif, derivative}
\usepackage{tikzsymbols}
\usepackage{flafter}
\usepackage{placeins}
\usepackage{dblfloatfix} 
\usepackage[whole]{bxcjkjatype}
\usepackage{comment}
\usepackage{here}
\newtheorem{theorem}{Theorem}[section]
\newtheorem{proposition}{Proposition}[section]

\usepackage[ruled,vlined,linesnumbered]{algorithm2e}

\newcommand{\argmin}{\operatorname*{argmin}}
\newcommand{\argmax}{\operatorname*{argmax}}

\makeatletter
\newcommand\vb{\@ifstar\mathbf\boldsymbol}
\newcommand\va[1]{\@ifstar{\vec{\mathrm{#1}}}{\vec{#1}}}
\newcommand\vu[1]{%
\@ifstar{\hat{\mathbf{#1}}}{\hat{\boldsymbol{#1}}}}
\makeatother
\newcommand{\D}{\text{D}}
\newcommand{\E}{\mathbb{E}}
\newcommand{\norm}[1]{\left\| #1 \right\|}
\newcommand{\R}{\mathbb{R}}

\newtheorem{result}{Result}

\title{The Effect of Optimal Self-Distillation\\in Noisy Gaussian Mixture Model}

\author{%
  Kaito Takanami\\
  Department of Physics\\
  Graduate School of Science, The University of Tokyo, \\
  Tokyo, Japan\\
  Center for Interdisciplinary AI and Data Science, Ochanomizu University\\
  Tokyo, Japan \\
  \texttt{takanami255@g.ecc.u-tokyo.ac.jp} \\
  \And
  Takashi Takahashi \\
  Institute for Physics of Intelligence, The University of Tokyo\\
  Tokyo, Japan\\
  RIKEN center for AIP\\
  \AND
  Ayaka Sakata \\
  Department of Information Science, Ochanomizu University\\
  Tokyo, Japan \\
  RIKEN center for AIP\\
}

\begin{document}

\maketitle

\begin{abstract}
Self-distillation (SD), a technique where a model improves itself using its own predictions, has attracted attention as a simple yet powerful approach in machine learning. Despite its widespread use, the mechanisms underlying its effectiveness remain unclear. In this study, we investigate the efficacy of hyperparameter-tuned multi-stage SD with a linear classifier for binary classification on noisy Gaussian mixture data. For the analysis, we employ the replica method from statistical physics. 
Our findings reveal that the primary driver of SD's performance improvement is denoising through hard pseudo-labels, namely discrete labels generated from the model’s own predictions, with the most notable gains observed in moderately sized datasets.
We also identify two practical heuristics to enhance SD: early stopping that limits the number of stages, which is broadly effective, and bias parameter fixing, which helps under label imbalance. To empirically validate our theoretical findings derived from our toy model, we conduct additional experiments on CIFAR-10 classification using pretrained ResNet backbone. These results provide both theoretical and practical insights, advancing our understanding and application of SD in noisy settings.
\end{abstract}

\section{Introduction}
\label{introduction}
Knowledge distillation (KD)~\cite{Hinton2015-es} is a technique in machine learning that transfers the learned information from a complex model (often referred to as the teacher) to a simpler model (the student). 
This method attracted attention for achieving model compression with minimal performance loss, and has been applied across various domains, including image classification~\cite{Liu2018-ib, Xu2020-am}, object detection~\cite{Chen2017-ty}, and natural language processing~\cite{calderon-etal-2023-systematic, Gu2023-pl}.

Among the various forms of KD, \emph{self-distillation} (SD), originally termed \emph{born again neural network}~\cite{Furlanello2018-vf} is particularly intriguing.
In SD, the teacher and student models share identical architectures. 
This means that SD does not attempt the model compression; rather, it retrains the student model using the teacher's output.
SD presents a intriguing paradox: despite training an identical model on the same dataset,
the student model can outperform the teacher~\cite{Furlanello2018-vf, hahn-choi-2019-self, clark-etal-2019-bam, yang-etal-2024-self, chen2025visiontransformersselfdistilledregisters}. 

Two main hypotheses have been proposed to explain such seemingly puzzling performance gains. 
The first suggests that the soft labels generated by the teacher provide \emph{dark knowledge}~\cite{Hinton2015-es}.
Here, dark knowledge refers to the information implicitly embedded in the prediction probability distribution of the teacher model's output, which is absent in hard labels. 
It provides the student with additional information that captures subtle relationships within the data.
The second hypothesis attributes the improvement to a denoising effect ~\cite{Das2023-zx, pmlr-v267-das25b} where the teacher model reduces the influence of the incorrect noisy labels in the training data, enabling the student model to learn a more reliable representation of the underlying patterns~\cite{Pareek2024-qf}.

Although these hypotheses offer plausible explanations, the optimal behavior of SD, achieved through hyperparameter optimization and repeated iterations~\cite{Pareek2024-qf}, remains poorly understood. This lack of understanding makes it difficult to identify the key factors that genuinely contribute to the performance improvement of SD. One reason for this difficulty is that exhaustive exploration of the hyperparameter space is usually computationally expensive, limiting the scope of experimental studies. As a result, evaluating the effectiveness of SD and identifying optimal strategies for its application remains a challenge.

To address this issue, we consider a multi-stage SD procedure using a linear classifier on Gaussian mixture data with label noise. This setup provides a controlled environment for analyzing both the dark knowledge and denoising hypotheses within a unified theoretical framework. In particular, we analyze this setting in the proportional asymptotic limit, where the input dimension $N$ and the data size $M$ diverge at the same rate, i.e., $N, M \to \infty$ with $M/N \to \alpha \in (0, \infty)$.
A salient feature of this proportional asymptotic regime is that it allows precise characterization of the trained classifier's behavior, rather than merely providing rough lower or upper bounds. This enables us to explicitly determine optimal hyperparameters and iteration procedures, at least within simplified settings. 
Moreover, because this precise characterization involves only a finite number of variables, an exhaustive search for the optimal hyperparameters becomes computationally feasible.
In this context, Gaussian mixture classification with linear models has served as a standard setting for gaining valuable insights into high-dimensional learning problems \cite{Dobriban2018high, Mignacco2020-mc, kini2020analytic, loureiro2021learning, deng2022model, mannelli2022unfair, pesce2023are, takahashi2024a}. 
Technical tools for analyzing such asymptotic regimes include the replica method~\cite{doi:10.1142/0271,
doi:10.1142/13341}, Convex Gaussian Min-max Theorem~\cite{Thrampoulidis2015-ex}, Approximate Message Passing~\cite{Donoho2009-lr}, which builds on Gordon's inequality~\cite{Gordon1988-mn}.
In our study of multi-stage SD, we employ the replica method, which has recently been shown to be applicable to multistage optimization problems, including self-training~\cite{Takahashi2022-qj} and alternating minimization~\cite{Okajima2024-ba}.

Our main results are as follows:
\begin{enumerate}
    \item
    The statistical properties of the trained classifiers are precisely characterized in the asymptotic limit where the input dimension and the data size diverge at the same rate. The precise formula for the generalization error is also be derived (Section \ref{sec:theory}).
    \item
    SD using soft labels with dark knowledge can outperform hard-label training, particularly under low-noise or limited-data conditions. However, the performance gains achieved with soft labels are often quantitatively comparable to those obtained using hard labels across all settings we investigated. These findings suggest that, at least within our toy model, dark knowledge is not the key factor driving the success of SD (Section~\ref{sec:soft_labels}).
    \item
    Naively applying multi-stage SD over too many stages degrades performance. This can be mitigated by employing an early-stopping heuristic that terminates the SD process at an appropriate stage. The resulting performance improvement is most pronounced for medium-sized datasets, where the denoising effect of SD is strongest. In addition, even though pseudo-labels contain label noise, performance comparable to training with ground-truth labels can be achieved in large-scale datasets (Section \ref{sec:t_stage}).
    \item
    When ground-truth labels are imbalanced, learning solely from the teacher's pseudo-labels becomes challenging. This difficulty arises because the optimal regularization strength for aligning decision boundaries differs from that required for appropriately estimating the bias term. It is shown that fixing the bias term in the early stages of multi-stage SD serves as an effective heuristic to mitigate this issue (Section~\ref{sec:bias}).
    \item
     Experiments on CIFAR-10 with a pre-trained ResNet backbone qualitatively validate several theoretical predictions, extending beyond toy-model settings (Section~\ref{sec:experiments}).
\end{enumerate}

These results provide a comprehensive understanding of the mechanisms underlying SD with a linear classifier on noisy Gaussian mixture data, and offer insights into how to optimally apply SD.

\textbf{Reproducibility: }
The codes to reproduce some of our results are available at \url{https://github.com/taka255/self-distillation-analysis}.

\textbf{Impact statement:} We believe this work, which is a theoretical study of the learning behavior of simple linear model in a synthetic setting, does not have notable societal consequences.


\section{Related Work} \label{sec:related_work}
\textbf{Replica method for multi-stage learning.}
The application of the replica method to analyzing the dynamics of high-dimensional systems was originally proposed for studying discrete optimization~\cite{Krzakala2007-nu} and stochastic processes in glassy systems~\cite{Franz2013-gy}. In recent years, it has been extended to learning problems, particularly for analyzing sequential optimization processes, including self-training in semi-supervised learning~\cite{Takahashi2022-qj} and alternating minimization ~\cite{Okajima2024-ba}. Our work builds on and advances this approach to analyze modern machine learning algorithms.

This methodology can be interpreted as a variant of Dynamical Mean Field Theory, a fundamental tool of statistical physics for analyzing the dynamics of high-dimensional systems, including gradient-based learning dynamics of neural networks~\cite{Zou2024-ms, Helias2020-kg} (see \cref{further_related_works} for details).

\textbf{Theoretical analysis of self-distillation.}
Theoretical analyses of distillation have predominantly focused on separable datasets due to their analytical tractability~\cite{Phuong2021-qh, pmlr-v267-das25b, Das2023-zx}. In separable settings, pseudo-labels generated by a teacher may be able to, in principle, match the ground-truth labels exactly. However, in realistic scenarios where data are not perfectly separable, pseudo-labels inevitably include errors. This fundamental limitation is not captured in these existing theoretical analyses, which rely on the separability assumption. While some exceptions~\cite{Ji2020-ws, Saglietti2022-do} have extended the analysis to non-separable datasets, they fall short of characterizing the behavior of optimal distillation under label noise. Our study fills this gap by quantitatively analyzing the improvements through SD with hyperparameter optimization on noisy and non-separable datasets.

\section{Notations and Problem Setup} \label{sec:problem_setup}
For convenience, we summarize all symbols and their definitions in Table~\ref{tab:notation} of Appendix~\ref{appendix:notation}, and provide a graphical illustration of the multi-stage SD model in Figure~\ref{fig:sketch} of the same appendix.

\subsection{Gaussian Mixture Data with Noisy Labels}
We consider the binary classification of Gaussian mixture data with noisy labels using a single-layer neural network. 
Let $\vb{x}_\mu \in \mathbb{R}^N$ be the input data, where $\mu = 1, \ldots, M$ is the index of the data point and $N$ is the dimension of the input data.
Here, we define the data-to-dimension ratio as $\alpha = M/N$.
The true labels $y_\mu^{\text{true}} \in \{0,1\}$ are independently generated according to the Bernoulli distribution $p(y_\mu^{\text{true}})=\rho^{y_\mu^{\text{true}}}(1-\rho)^{1-y_\mu^{\text{true}}}$, with $\rho\in(0,0.5]$.
We consider a noisy observation in which the observed labels $y_\mu\in\{0,1\}$ differ from the true labels with probability $\theta$: $\mathrm{Pr}[y_\mu\neq y_\mu^{\text{true}}]=\theta\in[0,1/2]$.
The feature vectors $\{\vb{x}_\mu\}_{1\le \mu \le M}$ are generated from the Gaussian mixture distribution:
\begin{equation}
  \vb{x}_\mu = (2y_\mu^{\text{true}} - 1) {\vb{v}}/{\sqrt{N}} + \sqrt{\Delta} \vb{z}_\mu, \label{eq:gaussian_mixture}
\end{equation}
where $\pm \vb{v} \in \mathbb{R}^N$ are the mean vectors of the Gaussian mixture, $\{\vb{z}_\mu\}_{1\le \mu \le M}$ are i.i.d. standard Gaussian vectors, 
and $\Delta > 0$ controls the variance of the additive noise.
\footnote{The results remain valid if $\vb{z}_\mu$ are replaced by i.i.d. random vectors with zero mean, unit covariance, and finite higher-order moments due to the central limit theorem.}.
Since the noise is rotation invariant, in the following, we set $\vb{v} = (1, 1, \ldots, 1)^\top$ without loss of generality.
The goal is to train a good classifier from $D_{\rm tr}=\{\bm{x}_\mu, y_\mu\}_{\mu=1}^M$ that can classify an unseen observation $\bm{x}$, generated in the same way as in \eqref{eq:gaussian_mixture}, correctly as $y^{\rm true}$.

\subsection{Multi-stage Self-Distillation Model}

We define the multi-stage SD model as a learning process that progresses through stages $t=0,1,2,\ldots$. The loss function at the $t$-th stage is given by
\begin{equation}
    \mathcal{L}_{t}(\vb{w}^{t}, B^{t}) = \sum_{\mu=1}^M \ell(y_\mu^t, Y(\vb{w}^t, B^t; \vb{x}_\mu)) + \frac{\lambda^{t}}{2} \norm{\vb{w}^{t}}^2 ,  \label{eq:loss_t}
\end{equation}
where $\ell(y,\hat{y})$ is a loss function, and the minimizer of Eq.~\eqref{eq:loss_t} is denoted as $\hat{\vb{w}}^t$ and $\hat{B}^t$.
Here, $Y(\vb{w}^t, B^t; \vb{x}_\mu)$ is the activation, and 
$y_\mu^t$ is the target label used for 
$t$-th stage learning. 
For $t = 0$, $y_\mu^t$ represents
the observed label $y_\mu$, and for $t > 0$, it is interpreted as the pseudo-label. The activations and pseudo-labels are defined by the following rules.

\textbf{Activations:}
The activation at the $t$-th stage is given by
\begin{equation}
  Y(\vb{w}^t, B^t; \vb{x}_\mu) = \sigma\ab(\frac{\vb{w}^t \cdot \vb{x}_\mu}{{\sqrt{N}}} + B^t), \label{eq:prediction}
\end{equation}
where $\sigma(x)$ is an activation function, and the factor $1/\sqrt{N}$ ensures the output remains at order $\mathcal{O}(1)$ in $N$. 
We use two loss–activation combinations: (i) cross-entropy loss and sigmoid activation, 
$\ell(y, \hat{y}) = -y \log \hat{y} - (1-y) \log (1-\hat{y})$, $\sigma(x) = 1/(1+\exp(-x))$;
(ii) mean squared error loss and linear activation, $\ell(y, \hat{y}) = (y - \hat{y})^2$, $\sigma(x) = (x+1)/2$. See Appendix~\ref{append:activation_function} for more details.
These choices ensure convexity of the loss at each stage, which is crucial for our theoretical analysis.
We refer to the model based on Eq.~\eqref{eq:loss_t} as the $t$-SD model, with the 0-SD model as the base model before distillation.
We refer to the $t$-SD model with cross-entropy loss as the \emph{logistic $t$-SD model}, and the version with mean squared error loss as the \emph{linear $t$-SD model}.

\textbf{Pseudo-Labels:} Labels at stage $t$ are computed from the $(t-1)$-th stage as
\begin{equation}
  {y}_\mu^t =   \sigma \ab(\beta^t \ab(\frac{{\hat{{\boldsymbol{w}}}^{t-1} \cdot \boldsymbol{x}^\mu}}{{\sqrt{N}}} + \hat{B}^{t-1})), \quad t \ge 1 \label{eq:soft_label},
\end{equation}
with $y_\mu^0 = y_\mu$. Here, $\beta^t > 0$ is the inverse temperature controlling the hardness of the pseudo-label. In the limit $\beta^t \to \infty$, it becomes hard ($0$ or $1$), whereas finite $\beta^t$ yields soft labels in $(0,1)$.
Note that this parameter is meaningful only in the logistic SD model, since the linear SD model is scale-invariant.

\subsection{Effect of Self-Distillation}
We evaluate the effect of SD by measuring generalization error with optimal hyperparameters, tuned to minimize it.
We define $\Theta^t=\{\lambda^i\}_{i=0}^t \cup \{\beta^j\}_{j=1}^t$ for $t \ge 0$, with $\Theta^0 = \{\lambda^0\}$, and define the error metrics:
\begin{align}
  \mathcal{E}^{t} &= \mathbb{E}_{\mathcal{D}} \mathbb{E}_{\vb{x}, y^\text{true}} \left[ \mathbb{I}\ab(\hat{Y}(\hat{\vb{w}}^t , \hat{B}^t; \vb{x}) \neq y^{\text{true}}) \right] \\
  \mathcal{E}^{*t} &= \min_{\Theta^t}  \mathcal{E}^{t}, \quad
  \mathcal{E}^{*t}_\text{Hard} = \min_{\lambda^{0}, \ldots, \lambda^t}  \lim_{\beta^1, \cdots, \beta^t \to \infty}  \mathcal{E}^t \quad (t\geq 1) \label{eq:hard_e},
\end{align}
where $\hat{Y}(\hat{\vb{w}}^t, \hat{B}^t; \vb{x})=\mathbb{I}[Y(\hat{\bm{w}}^t, \hat{B}^t;\bm{x})>1/2]$, 
$\mathbb{I}(A)$ is the indicator function, $\mathcal{D}= \{(\vb{x}_\mu, y_\mu^\text{true}, y_\mu)\}_{1\le \mu \le M}$, and $(\bm{x}, y^{\mathrm{true}})$ is a test input generated in the same way as the training data.
$\mathcal{E}^{*t}_\text{Hard}$ represents the error when dark knowledge is removed by hardening soft labels.
This limit is meaningful only for the logistic $t$-SD model.
We include this as a reference to assess the role of dark knowledge.
A significantly smaller $\mathcal{E}^{*t}$ than $\mathcal{E}^{*0}$ indicates SD effectively improves generalization.

\section{Precise characterization of multi-stage self-distillation} \label{sec:theory}

For a new input $\bm{x}$ generated from the Gaussian mixture as in \eqref{eq:gaussian_mixture}, the distribution of the pre-activation $\hat{\bm{w}}^t\cdot \bm{x}/\sqrt{N} + \hat{B}^t$ can be characterized as 
\begin{equation}
    \frac{\hat{\bm{w}}^t\cdot \bm{x}}{\sqrt{N}} + \hat{B}^t \overset{\rm d}{=} {\bar{m}^t}(2y^{\rm true}-1) + \hat{B}^t + \sqrt{\Delta \bar{Q}^{tt}}z,
\end{equation}
where $\overset{\mathrm{d}}{=}$ denotes equality in distribution, $\bar{Q}^{tt} = \|\hat{\bm{w}}^t\|^2/N$ represents the norm of the weight vector, $\bar{m}^t = \hat{\bm{w}}^t \cdot \bm{v}/N$ is its alignment with the cluster center direction $\bm{v}$, and $z \sim \mathcal{N}(0,1)$. The label $y^{\rm true}$ follows $p(y^{\rm true}) = \rho^{y^{\rm true}} (1 - \rho)^{1 - y^{\rm true}}$.
As such, the term $\bar{m}^t(2y^{\rm true}-1)$ represents the signal component, while $\sqrt{\bar{Q}^{tt}}$ controls the uncertainty due to the variance in the classifier's weights.
This result follows from the independence of $\bm{x}$ from $D_{\rm tr}$ and its Gaussianity. 

At the proportional limit $N,M\to\infty, M/N\to\alpha\in(0,\infty)$, these quantities are expected to converge to deterministic values, which do not fluctuate aginst the realization of $D_{\rm tr}$, as 
\begin{equation}
    \bar{Q}^{tt}(D_{\rm tr}) \to Q^{tt}, 
    \quad 
    \bar{m}^t(D_{\rm tr})\to m^t,
    \quad
    \hat{B}^t(D_{\rm tr}) \to b^t.
    \label{eq:key quantities}
\end{equation}
This leads to the following expression of the average generalization error \cite{Mignacco2020-mc}.

\begin{proposition}
    \label{prop:generalization}
    Under the proportional asymptotic limit ($N, M\to\infty$, constrained by $M/N \to \alpha \in(0,\infty)$), 
    the average generalization error of the $t$-SD model is given by
    \begin{equation}
        \mathcal{E}^t = \rho H\ab(\frac{m^t + b^t}{\sqrt{\Delta Q^{tt}}}) + (1-\rho) H\ab(\frac{m^t - b^t}{\sqrt{\Delta Q^{tt}}}) , \label{eq:generalization_error}
    \end{equation}
    where $H(x) = 1-  \int_{-\infty}^x \d t \, e^{-t^2/2} /\sqrt{2\pi} $.
\end{proposition}

Proposition~\ref{prop:generalization} indicates that the key quantities to evaluate the generalization error are the \emph{alignment} $\bar{m}^t/\sqrt{\bar{Q}^{tt}}$ (the cosine similarity between the direction of the decision boundary and $\vb{v}$) and the \emph{rescaled bias} $\hat{B}^t/\sqrt{\bar{Q}^{tt}}$ (the offset of the decision boundary from the origin).

Since $\bar{Q}^{tt}, \bar{m}^t$ and $\hat{B}^t$ are expected to be concentrated to deterministic values as in \eqref{eq:key quantities}, they are evaluated by investigating the average values $\E_{\mathcal{D}}[\bar{Q}^{tt}], \E_{\mathcal{D}}[\bar{m}^t], \E_{\mathcal{D}}[\hat{B}^t]$ at the large system limit. For evaluating these averaged quantities, we used the replica method and obtained the following results, which precisely characterize the values of them.

\begin{result}[\textbf{Statistics of the T-SD model parameters}]
    \label{thm:replica}
    There exist constant matrices
    $\hat{Q} = (\hat{Q}^{st}), \hat{\chi} = (\hat{\chi}^{st})\in\mathbb{R}^{(T+1)\times(T+1)}$
    and a constant vector
    $\hat{m}=(\hat{m}^t)\in\mathbb{R}^{T+1}$
    such that, in the proportional asymptotic limit
    ($N, M\to\infty, M/N \to \alpha \in(0,\infty)$), 
    \begin{equation}
        \hat{w}^0_i \overset{\mathrm{d}}{=} \frac{1}{\hat{Q}^{00} + \lambda^0}  \ab(\hat{m}^0 + \hat{\xi}^0)  \quad \text{and} \quad \hat{w}^T_i \overset{\mathrm{d}}{=} \frac{1}{\hat{Q}^{TT} + \lambda^T}  \ab(\hat{m}^T + \hat{\xi}^T - \sum_{s=0}^{T-1}\hat{Q}^{st}\hat{w}^s) \quad (T\geq 1) \label{eq:wt_main},
    \end{equation}
    where $\hat{\vb{\xi}}=(\hat{\xi}^t)\in\R^{T+1} \sim \mathcal{N}(\vb{0}, \hat{\chi})$.
\end{result}
We deliberately use the notation \emph{Result} rather than \emph{Theorem}, as the derivation relies on the replica method is not yet a mathematically rigorous proof.\footnote{For possible directions toward 
rigorous proofs and the associated technical challenges, see the remark in \cref{sec:remarks_on_rigorous}.}. 
The derivation is given in \cref{append:replica}, where the explicit determination of the constant matrices $\hat Q, \hat \chi$ and the vector $\hat m$ is also detailed. 
Furthermore, Result~\ref{thm:replica} is validated by numerical simulations in \cref{append:agreement}, demonstrating good consistency.

The constants $\hat{Q},\hat{m}$ and $\hat{\chi}$ that appear in Theorem~\ref{thm:replica} and $b^t$ in Equation~\ref{eq:generalization_error} can all be obtained by solving at most $\mathcal{O}(T)$ coupled equations (see Appendix~\ref{append:replica}), so they can be computed efficiently.  
Moreover, from Equation~\eqref{eq:wt_main} one directly obtains $Q^{tt}=\mathbb{E}[(\hat{w}_i^t)^2]$ and $m^{t}=\mathbb{E}[\hat{w}_i^t]$. 
Together with Proposition~\ref{prop:generalization}, this gives an exact characterization of the generalization error in the asymptotic limit.

Beyond yielding a closed‐form for the asymptotic generalization error, Theorem~\ref{thm:replica} also makes clear that the parameters learned via SD are effectively governed by only a small number of key order parameters. To see how this plays out in practice, we now unpack the roles of the parameters in Theorem~\ref{thm:replica}.
The weights are composed of three components: 
(i) the signal term $\hat{m}^t$ 
representing the amount of correlation with the cluster center $\bm{v}$,
(ii) the noise term $\hat{\xi}^t$ capturing the randomness inherent in the data, and 
(iii) the correction term $-\sum_{s=1}^{t-1}\hat{Q}^{st}\hat{w}^s$
accounting for correlations induced by the labels generated 
by previous teacher models, i.e. $0$-th to $(t-1)$-th SD models.

\section{The role of soft labels in self-distillation} \label{sec:soft_labels}

In this section, we focus on the $t=1$ case, where SD involves a single teacher and student.

\begin{figure}[htb]
    \centering
    \includegraphics[width=0.35\columnwidth]{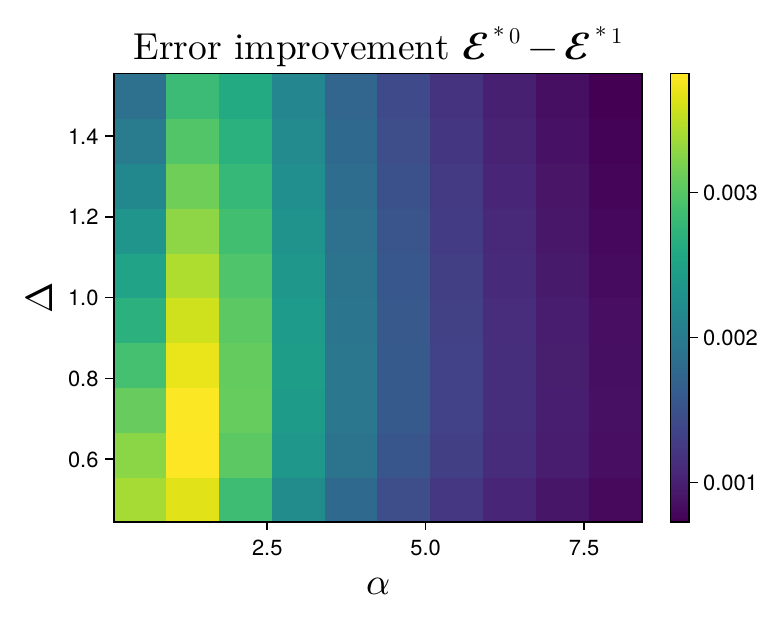}
    \vskip -0.1in
    \caption{
    Heat map of the improvement error $\mathcal{E}^{*0} - \mathcal{E}^{*1}$ at $\rho=0.4$ and $\theta=0$ in linear $1$-SD model.}
    \label{fig:noiseless}
\end{figure}

\textbf{Dark knowledge effect is marginal.}
We investigate the generalization error improvement of the optimal logistic $1$-SD model in a noiseless setting ($\theta=0$), where any improvement would stem solely from the teacher's soft labels, because there is no label noise to be removed. 
As shown in Figure~\ref{fig:noiseless}, the linear $1$-SD model exhibits only modest improvement across various dataset sizes and data variances. 
Gains are slightly more visible when the dataset is small and variance is low, but even under such conditions, the improvement remains under $0.4\%$.
These results suggest that in linear models, the contribution of dark knowledge is limited.

\textbf{Soft labels vs. hard labels.}
Soft labels provide dark knowledge beyond simple class predictions. 
However, since denoising primarily aims to identify the correct label, this additional information may not always be necessary.
Motivated by the limited effect of dark knowledge observed in the noiseless case, we hypothesize that hard labels, obtained by binarizing soft labels, may suffice for SD.
To test this hypothesis, we compare generalization errors in logistic $1$-SD using hard and soft labels.

\begin{figure*}[htb]
    \centering
    \includegraphics[width=\columnwidth]{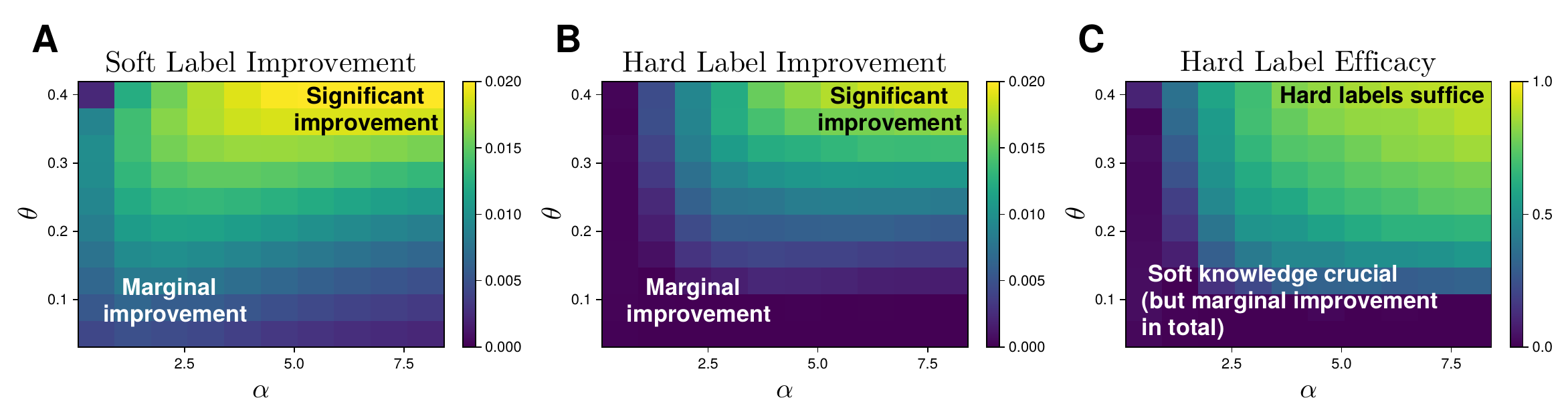}
    \vskip -0.1in
    \caption{
        (A) and (B): generalization error improvements in the optimal logistic 1-SD model using soft labels $(\mathcal{E}^{*0} - \mathcal{E}^{*1})$ and hard labels $(\mathcal{E}^{*0} - \mathcal{E}^{*1}_\text{Hard})$, respectively. 
        (C): the ratio of the two: $(\mathcal{E}^{*0} - \mathcal{E}^{*1}_\text{Hard}) / (\mathcal{E}^{*0} - \mathcal{E}^{*1})$. 
        Parameters: $\rho = 0.4$, $\Delta = 1.0$.
        }
    \label{fig:2stage_heatmap}
\end{figure*}

Figure~\ref{fig:2stage_heatmap}A and B show the improvement in generalization errors achieved using soft labels ($\mathcal{E}^{\ast0} - \mathcal{E}^{\ast 1}$) and hard labels ($\mathcal{E}^{\ast0} - \mathcal{E}^{\ast1}_\text{Hard}$), respectively. 
Both exhibit similar qualitative trends: large improvements are observed in higher noise and larger dataset settings, consistent with the denoising interpretation. 

Figure~\ref{fig:2stage_heatmap}C shows the ratio of the improvements in A and B, revealing two distinct regions. In the dark purple region, soft labels offer a clear advantage over hard labels, indicating that dark knowledge contributes meaningfully to SD. In contrast, the bright yellow region shows that hard labels are nearly as effective as soft labels, suggesting that dark knowledge is unnecessary in these conditions. However, even in the region where soft labels outperform hard labels, the quantitative improvement is small (Figure~\ref{fig:2stage_heatmap}A). These findings suggest that while soft labels may be beneficial in specific conditions, their overall impact on SD performance is relatively limited in our setting.

This observation may refine our understanding of dark knowledge in SD. While previous studies~\cite{Ma2022-ij, pmlr-v283-mandal25a} emphasized its general importance, our results suggest that its effectiveness depends on dataset characteristics and noise level, and that it is not always necessary.

\section{Understanding the effect of multi-stages} \label{sec:t_stage}
So far, we have examined the role of dark knowledge using the logistic $t$-SD model in single-stage settings. Here, we shift our focus and consider the effect of repeated SD across multiple stages.

To investigate this, we consider the $t$-stage model ($t \geq 2$) and study how performance evolves with repeated applications of SD. Due to computational constraints, we focus on the linear $t$-SD model. Although it does not distinguish between soft and hard labels, we expect it to still capture the essential behavior of multi-stage SD.
We first focus on the label-balanced case ($\rho = 0.5$), and discuss the label-imbalanced setting ($\rho < 0.5$) in Section~\ref{sec:bias}.

\begin{figure*}[htb]
    \centering
    \includegraphics[width=1.0\columnwidth]{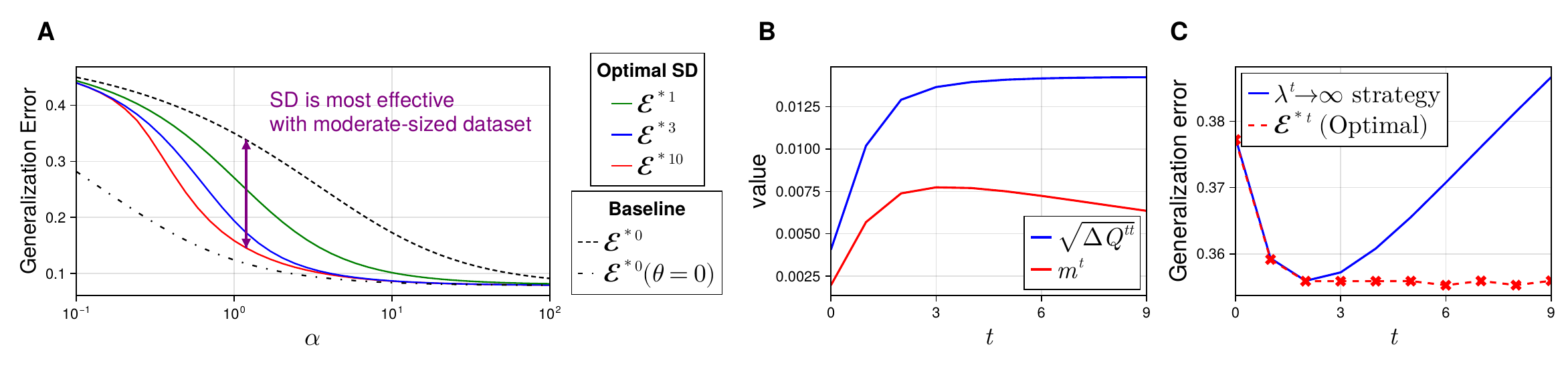}
    \vskip -0.1in
    \caption{(A) A comparison of the optimal generalization error for the linear $t$-SD model, $0$-SD model, and the noiseless case.
    (B) Dynamics of $\sqrt{\Delta Q^{tt}}$ and $m^t$ for the linear $t$-SD model with $\lambda^0, \cdots, \lambda^t \to\infty$.  
    (C) Comparison of generalization error between the linear $t$-SD model with $\lambda^0, \cdots, \lambda^t \to\infty$ and optimal $t$-SD. 
    Parameters for (A): $\rho = 0.5, \Delta = 0.5, \theta = 0.4$. (B, C): $\alpha = 1.0, \Delta = 1.2, \theta = 0.3, \beta^t = 1/\sqrt{Q^{t-1,t-1}}$.}
    \label{fig:naive_strategy}
\end{figure*}

\textbf{Denoising effects and dataset size dependence.}
We investigated the denoising effect of SD by comparing the optimal $t$-SD model with two baselines: the optimal $0$-SD model ($\mathcal{E}^{*0}$ with $\theta>0$) and the optimal $0$-SD model trained on noiseless data ($\mathcal{E}^{*0}$ with $\theta=0$), as shown in Figure~\ref{fig:naive_strategy}A.

The behavior of ${\cal E}^{*t}$ can be categorized into three types depending on $\alpha$: large $\alpha$, intermediate $\alpha$, and small $\alpha$. When $\alpha$ is sufficiently large ($\alpha\gtrsim10$) or small ($\alpha \lesssim 0.2$), the decrease in ${\cal E}^{*t}$ slows down around $t=3$, with ${\cal E}^{*3}$ and ${\cal E}^{*10}$ being almost the same, as shown in Figure~\ref{fig:naive_strategy}A. 

At sufficiently large $\alpha$ ($\alpha\gtrsim10^1$ in Figure \ref{fig:naive_strategy}A), multi-stage SD quickly achieves performance close to that of noise-free setting. Specifically, by $t = 3$, the generalization error ${\cal E}^{*t}$ becomes nearly equal to the noise-free case ${\cal E}^{*0}$ with $\theta=0$, even though the error before distillation ${\cal E}^{*0}$ with $\theta>0$ is clearly higher than noise-free case.
This finding is noteworthy, as perfect noise correction is challenging due to the overlapping data distributions.
In contrast, for small $\alpha$, $\mathcal{E}^{*t}$ remains close to $\mathcal{E}^{*0}$ even at $t=10$, indicating that SD's denoising effect is limited.
Unlike large or small $\alpha$, in the intermediate range, the improvement progresses slowly with the number of iterations,
but ${\cal E}^{*t}$ approaches to the noise-free case as $t$ increases.
As a consequence, the error improvement by multi-stage SD exhibits a non-monotonic behavior with dataset size, with the most pronounced denoising effect observed for moderately-sized datasets, as indicated by the arrow in Figure \ref{fig:naive_strategy}A.

Intuitively, this phenomenon can be explained as follows. For small datasets, the limited data makes it difficult for the teacher to learn pseudo-labels that effectively correct label noise, reducing the effectiveness of SD. In contrast, for large datasets, the teacher becomes strong enough to generate pseudo-labels that allow SD to correct noise effectively. However, the pre-SD classifier already performs well, leaving little room for further improvement. For intermediate dataset sizes, the available data is sufficient for the teacher to produce pseudo-labels that enable noise correction. At the same time, the pre-SD classifier remains suboptimal. As a result, the most substantial performance gains are observed in this intermediate $\alpha$ region.

\textbf{Fixed point analysis and learning dynamics.} A natural question may be how the error of $t$-SD behaves as $t \to \infty$. To investigate this, the limit $\lambda^0, \dots, \lambda^t \to \infty$ is considered, under which a closed-form expression for the generalization error of the linear $t$-SD model can be derived. This provides theoretical predictions for the error as $t \to \infty$. This setting is not only mathematically tractable but also interpretable: it simplifies the solution to a the averaging estimator $\hat{\bm{w}}^t \propto \sum_{\mu=1}^M y_\mu \bm{x}_\mu$. As shown in \cite{Dobriban2018high, Lelarge2019-gx, Mignacco2020-mc}, this averaging estimator is Bayes-optimal when noise-less ($\theta=0$) and balanced ($\rho=1/2$) case. 
Furthermore, \cite{10619631} shows that in the $\lambda \to \infty$ regime, any corruption rate ($\theta < 0.5$) in the balanced case ($\rho=1/2$) can be eliminated.
We also confirmed in preliminary experiments that the optimal hyperparameter schedule obtained by black-box optimization starts with very large regularization and then switches to zero, which naturally supports early stopping as a sensible strategy. 
Based on the above considerations, we now present the following theorem for this large regularization limit.
\begin{result}[\textbf{The generalization error at $t\to\infty$}]
    For an arbitrary choice of the set of the temperature parameters $\{\beta^t\}_{t\ge0}$,
    the generalization error of the linear $t$-SD model with $\rho=0.5$, $\lambda^0, \cdots, \lambda^t \to\infty$ and $t\to\infty$ is given by $\lim_{t\to\infty} \mathcal{E}^t = 0.5$ whenever $\alpha<\Delta^2$.
    \label{prop:fixedpoint}
\end{result}
\vskip-0.05in
The proof is given in \cref{append:exact_solution}.

Result~\ref{prop:fixedpoint} shows that, under certain data conditions, naively continuing multi-stage SD can reduce the model's performance to the level of random guessing. In particular, when $\alpha = \Delta^2$, the generalization error exhibits a phase transition that separates meaningful predictors from random ones (see Appendix~\ref{append:exact_solution}).

To gain deeper insights into the learning dynamics, we analyze the time evolution of $m^t = \mathbb{E}\ab[\hat{\bm{w}}^t \cdot \vb{v}] / N$ and $Q^{tt} = \mathbb{E}\ab[\hat{\bm{w}}^t \cdot \hat{\bm{w}}^t] / N$, which quantify signal extraction and prediction uncertainty, respectively. The temperature is set to $\beta^t = 1 / \sqrt{Q^{t-1,t-1}}$ to prevent the norm of the weight vector, $\|\hat{\bm{w}}^t\|$, from vanishing. The values of $m^t$ and $Q^{tt}$ are plotted in Figure~\ref{fig:naive_strategy}B. 
As shown in the figure, \(m^t\) peaks during the initial iterations, whereas the predictive uncertainty \(\sqrt{\Delta Q^{tt}}\) increases steadily throughout the stages, leading to performance decline.
Therefore, optimal learning may be achieved by halting the training process when signal extraction is maximized. 
Interestingly, this early stopping strategy closely matches the results obtained through comprehensive hyperparameter optimization across the entire model (Figure~\ref{fig:naive_strategy}C).

These results are consistent with experimental studies~\cite{Zhang2020-bh}, 
where the term ``diversity'' specifically refers to the predictive uncertainty of teacher predictions, which has been suggested to relate to the success of SD. 
Our result may support that such predictive uncertainty (diversity) plays a key role in effective signal extraction. However, the results also imply that the extractable signal saturates after a few iterations, highlighting an intrinsic limit to the benefit of repeated distillation.

\section{The hardness of learning bias in label imbalanced cases} \label{sec:bias}
Next, we examine the label imbalanced case $\rho < 0.5$, where performance improvement results from the interplay between alignment and rescaled bias, and compare it to the $\rho = 0.5$ case.

\begin{figure*}[htb]
    \centering
    \centerline{\includegraphics[width=1.\textwidth]{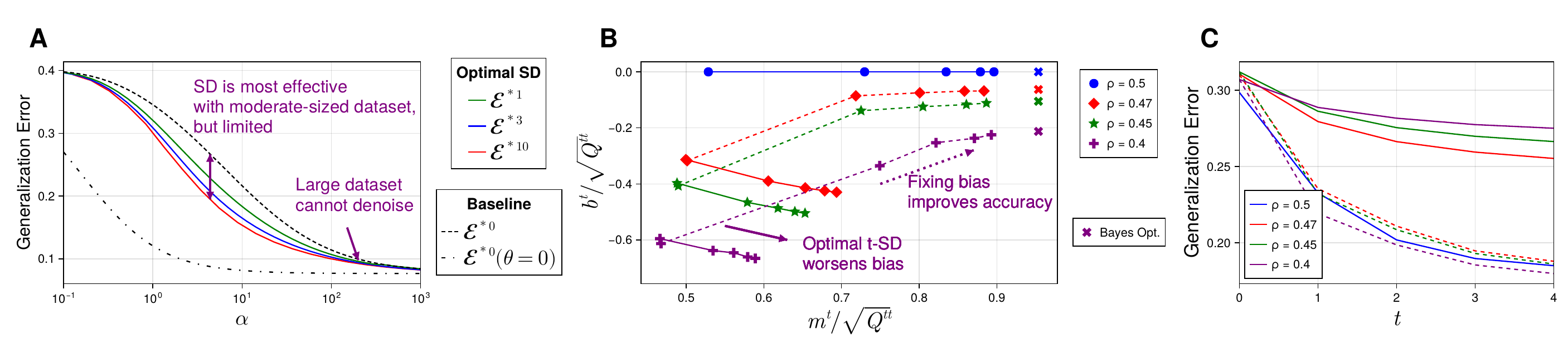}}
    \vskip -0.1in
    \caption{
    (A) Optimal generalization error of the linear $t$-SD model compared with the $0$-SD model and the noiseless case under label imbalance ($\rho = 0.4$). 
    (B) Evolution of the rescaled bias ($|b^t|/\sqrt{Q^{tt}}$) and alignment ($m^t/\sqrt{Q^{tt}}$) from $t = 0$ to $t = 4$ for the optimal $t$-SD model (solid lines) and the variant with fixed bias (dotted lines). 
    (C) Generalization error over stages $t$ for the same models as in (B). Parameters: (A) $\rho = 0.4$, $\Delta = 0.5$, $\theta = 0.4$; (B, C) $\Delta = 1.0$, $\theta = 0.4$, $\alpha = 10.0$.
    }
    \label{fig:bias_effect}
\end{figure*}

The difference from the label-balanced case lies in the difficulty of simultaneously learning the bias and the alignment in imbalanced datasets. Figure~\ref{fig:bias_effect}B shows the evolution of the rescaled bias ($b^t/\sqrt{Q^{tt}}$) and alignment ($m^t/\sqrt{Q^{tt}}$) over distillation stages. As the figure indicates, while alignment improves gradually, the rescaled bias worsens as training progresses, deviating from its Bayes-optimal value.

This behavior can be attributed to the effect of ridge regularization, which acts only on the weight vector $\vb{w}^t$. Strong regularization, which may be necessary to improve alignment, shrinks the norm the weight ($Q^{tt}=\|\hat{\vb{w}}^t\|_2^2/N$) and consequently increases the rescaled bias. When the rescaled bias becomes too large compared to the alignment, the model tends to classify most data into a single class (either positive or negative), resulting in poor generalization performance (see also Eq.~\eqref{eq:generalization_error}).

Hence, in label-imbalanced cases, loss minimization may not be suitable for jointly optimizing bias and weight. In contrast, for balanced data, the optimal bias is simply zero ($b^t = 0$), and no such trade-off arises.

To address the challenge of balancing bias and alignment, we find that fixing the bias at an early stage is a simple and effective heuristic. Similar approaches, which separate the training of alignment and bias, have also been proposed in logistic regression  \cite{Mignacco2020-mc} and self-training~\cite{Takahashi2022-qj} in imbalanced Gaussian mixtures. The dotted lines in Figure~\ref{fig:bias_effect}B illustrate the results with the bias fixed at its value obtained in the optimal $0$-SD, followed by performing the optimal $t$-SD. Figure~\ref{fig:bias_effect}C further compares the generalization error of $t$-SD with and without this heuristic. Applying bias fixing significantly improves both rescaled bias and alignment,  and exhibits convergence towards the Bayes-optimal solution, as observed in the $\rho = 0.5$ case.

\section{Experiments on real datasets}\label{sec:experiments}
We have analyzed the behavior of multi-stage SD using the Gaussian mixture model with label noise, which allowed precise asymptotic characterization and provided valuable insights. However, this setting is highly idealized, and there remains a significant gap between this toy model and real-world datasets. To bridge the gap, we conduct a sanity check to test whether our theoretical predictions hold in a standard vision task. We fine-tune only the final layer of a ResNet pretrained on IMAGENET1K\_V2~\cite{torchvision2016} (BSD 3-Clause ``New'' License) with $L_2$ regularization on noisy CIFAR-10 (cat vs.\ dog)~\cite{Krizhevsky2009-ac} (MIT License) and compare the results to our theoretical predictions.  See Appendix~\ref{append:experiment} for experimental details.

\begin{figure}[htb]
    \centering
    \includegraphics[width=0.8\textwidth]{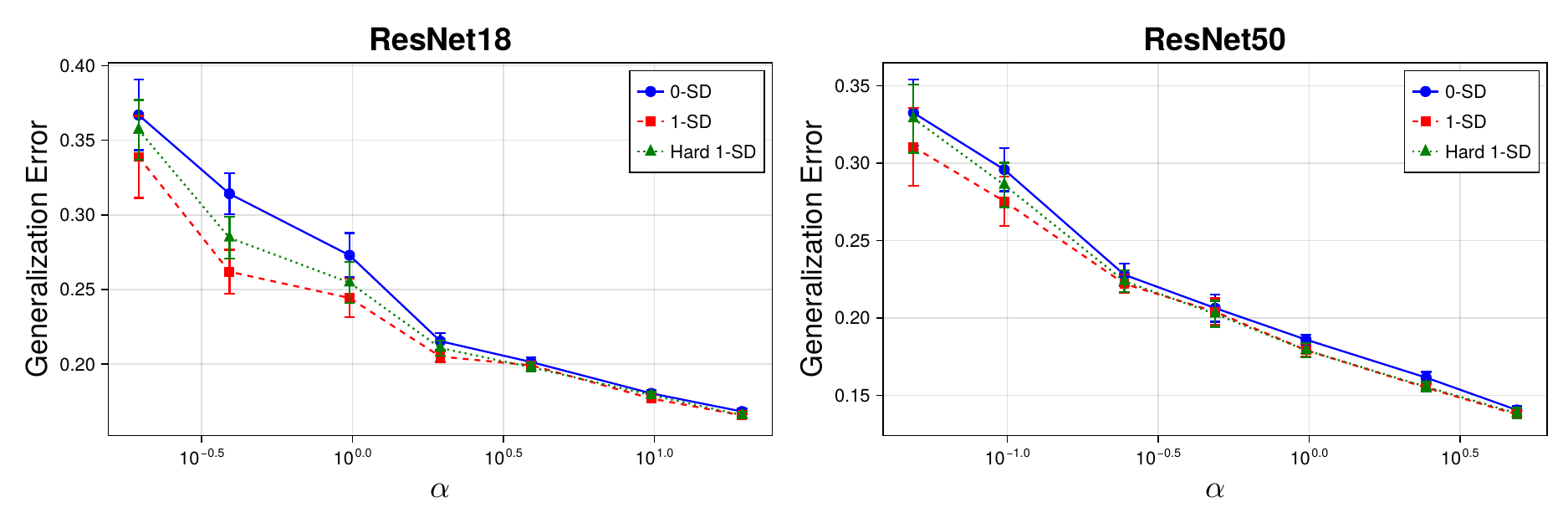}
    \caption{Comparison of the optimal generalization error of the logistic $0$-SD model, $1$-SD model and $1$-SD model using hard pseudo labels for CIFAR-10 dog versus cat classification using pretrained ResNet-18 ($N=512$) and ResNet-50 ($N=2048$) feature representations. Parameters: $\theta=0.4$.  Error bars represent the standard error of the mean over 10 trials per point.}
    \vskip-0.1in
    \label{fig:real_experiment_main}
    \vskip-0.05in
\end{figure}

Figure~\ref{fig:real_experiment_main} shows the generalization error of optimal $1$-SD, compared to optimal $0$-SD and optimal $1$-SD using hard labels (see Eq.~\ref{eq:hard_e}). 
When using ResNet-18 for feature extraction, we observe virtually no improvement due to SD in the large-$\alpha$ region, while in the middle-$\alpha$ region SD achieves a denoising effect that yields $\sim 5\%$ performance gain. According to the results in Section~\ref{sec:t_stage}, the benefit of SD should peak at moderate $\alpha$ and decline again as $\alpha$ becomes even smaller; however, this downturn cannot be observed here because it corresponds to an excessively small sample size in our setup. 
Nevertheless, Figure~\ref{fig:real_experiment_main} partially reproduce the prediction from Section~\ref{sec:t_stage} that SD's effectiveness is maximized at intermediate values of $\alpha$.

Next, focusing on the improvement in generalization error when propagating hard labels from $0$‐SD to $1$‐SD, we find that the relative importance of soft‐label error reduction grows as $\alpha$ decreases, mirroring the results of Section~\ref{sec:soft_labels}. 
Remarkably, as anticipated in Section~\ref{sec:soft_labels}, even hard labels, which completely discard dark knowledge, still provide a significant denoising benefit.
We observe a similar trend when using ResNet‐50; however, since this model has a higher baseline performance, the overall magnitude of improvement is smaller. See also Appendix~\ref{append:additional_exp} for further results.

\section{Conclusion}\label{sec:conclusion}
We investigated optimal multi-stage SD with a linear classifier for binary classification on noisy Gaussian mixture data. The technical crux of our analysis is a precise asymptotic formula for multi-stage SD, derived using the replica method from statistical physics. As this formula involves only a small, finite number of variables, it enables exhaustive hyperparameter search at a reasonable computational cost. This contrasts with experimental studies, where exhaustive exploration of the hyperparameter space is often computationally intractable.
By using this, we obtained the following results.
First, we found that dark knowledge in soft-labels plays a more limited role than previously assumed, with denoising likely being the primary driver of its success. SD's strong denoising capability is evident even with inseparable data distributions. 
Second, SD is most effective with moderate dataset sizes, showing weaker effects in both very small datasets (where denoising is difficult) and very large datasets (where noise has small impact). 
Third, fixing the bias and focusing on alignment optimization serves as a useful heuristic in SD. More broadly, this suggests a general strategy for multi-stage SD: progressively narrowing the parameters optimized at each stage. 
These findings not only enhance our theoretical understanding of SD mechanisms but also provide a foundation for developing improved algorithms. 

\textbf{Limitations: }
While our asymptotic analysis provides valuable insights, it is limited to linear models under a Gaussian mixture setting. Extending the analysis to deep networks or alternative SD strategies remains an important direction for future work. See Appendix~\ref{append:further_remarks} for further discussion.

\section*{Acknowledgements}
The study was conducted as part of the exploratory project ``Mathematical Exploration of Universal Structures in Multicomponent and Polydisperse Systems," supported by the Toyota Konpon Research Institute, Inc. This work was also supported by JSPS KAKENHI Grant Numbers 22H05117 and 23K16960, JST ACT-X Grant Number JPMJAX24CG, and JST BOOST NAIS Grant Number JPMJBS2418.

\bibliographystyle{unsrtnat}   
\bibliography{refs}

@ARTICLE{Mignacco2020-mc,
  title     = "The role of regularization in classification of high-dimensional
               noisy Gaussian mixture",
  author    = "Mignacco, Francesca and Krzakala, Florent and Lu, Yue M and
               Zdeborov'a, Lenka",
  editor    = "Iii, Hal Daum\'{e} and Singh, Aarti",
  journal   = "ICML",
  publisher = "PMLR",
  volume    =  119,
  pages     = "6874--6883",
  month     =  "26~" # feb,
  year      =  2020,
  url       = "https://proceedings.mlr.press/v119/mignacco20a.html",
  eprint    = "2002.11544"
}

@INPROCEEDINGS{Furlanello2018-vf,
  title     = "Born Again Neural Networks",
  author    = "Furlanello, Tommaso and Lipton, Zachary and Tschannen, Michael
               and Itti, Laurent and Anandkumar, Anima",
  editor    = "Dy, Jennifer and Krause, Andreas",
  booktitle = "Proceedings of the 35th International Conference on Machine
               Learning",
  publisher = "PMLR",
  volume    =  80,
  pages     = "1607--1616",
  series    = "Proceedings of Machine Learning Research",
  year      =  2018,
  url       = "https://proceedings.mlr.press/v80/furlanello18a.html"
}

@ARTICLE{Hinton2015-es,
  title         = "Distilling the knowledge in a neural network",
  author        = "Hinton, Geoffrey and Vinyals, Oriol and Dean, Jeff",
  journal       = "arXiv [stat.ML]",
  month         =  "9~" # mar,
  year          =  2015,
  url           = "http://arxiv.org/abs/1503.02531",
  archivePrefix = "arXiv",
  primaryClass  = "stat.ML",
  eprint        = "1503.02531"
}

@ARTICLE{Ma2022-ij,
  title    = "Sparse logits suffice to fail knowledge distillation",
  author   = "Ma, Haoyu and Huang, Yifan and Tang, Hao and You, Chenyu and Kong,
              Deying and Xie, Xiaohui",
  journal  = "ICLR 2022 Workshop on",
  month    =  "25~" # mar,
  year     =  2022,
  url      = "https://openreview.net/forum?id=BxZgduuNDl5"
}

@ARTICLE{Takahashi2022-qj,
  title         = "The role of pseudo-labels in self-training linear classifiers
                   on high-dimensional Gaussian mixture data",
  author        = "Takahashi, Takashi",
  journal       = "arXiv [stat.ML]",
  month         =  "16~" # may,
  year          =  2022,
  url           = "http://arxiv.org/abs/2205.07739",
  archivePrefix = "arXiv",
  primaryClass  = "stat.ML",
  eprint        = "2205.07739"
}

@ARTICLE{Barbier2019-wh,
  title     = "Optimal errors and phase transitions in high-dimensional
               generalized linear models",
  author    = "Barbier, Jean and Krzakala, Florent and Macris, Nicolas and
               Miolane, L\'{e}o and Zdeborov\'{a}, Lenka",
  journal   = "Proc. Natl. Acad. Sci. U. S. A.",
  publisher = "Proceedings of the National Academy of Sciences",
  volume    =  116,
  number    =  12,
  pages     = "5451--5460",
  month     =  "19~" # mar,
  year      =  2019,
  url       = "https://www.pnas.org/doi/abs/10.1073/pnas.1802705116",
  doi       = "10.1073/pnas.1802705116",
  pmc       = "PMC6431156",
  pmid      =  30824595,
  issn      = "0027-8424,1091-6490"
}

@ARTICLE{Gardner1988-lo,
  title     = "The space of interactions in neural network models",
  author    = "Gardner, E",
  journal   = "J. Phys. A Math. Gen.",
  publisher = "IOP Publishing",
  volume    =  21,
  number    =  1,
  pages     = "257--270",
  month     =  "7~" # jan,
  year      =  1988,
  url       = "https://iopscience.iop.org/article/10.1088/0305-4470/21/1/030/meta?casa_token=y0DToM4NETIAAAAA:OnP8Zv-34wfKOU5Bl1TYgXjJxf5xgcQdbVQ0doXLG2WFEB0DV0mC-y3-oG0iEc5A8c15w_1DYUkevVQQ--uCcqMPzsnK8w",
  doi       = "10.1088/0305-4470/21/1/030",
  issn      = "0305-4470,1361-6447"
}

@INPROCEEDINGS{Saglietti2022-do,
  title     = "Solvable Model for Inheriting the Regularization through
               Knowledge Distillation",
  author    = "Saglietti, Luca and Zdeborova, Lenka",
  editor    = "Bruna, Joan and Hesthaven, Jan and Zdeborova, Lenka",
  booktitle = "Proceedings of the 2nd Mathematical and Scientific Machine
               Learning Conference",
  publisher = "PMLR",
  volume    =  145,
  pages     = "809--846",
  series    = "Proceedings of Machine Learning Research",
  year      =  2022,
  url       = "https://proceedings.mlr.press/v145/saglietti22a.html"
}

@ARTICLE{Franz2013-gy,
  title     = "Quasi-equilibrium in glassy dynamics: an algebraic view",
  author    = "Franz, Silvio and Parisi, Giorgio",
  journal   = "J. Stat. Mech.",
  publisher = "IOP Publishing",
  volume    =  2013,
  number    =  02,
  pages     = "P02003",
  month     =  "1~" # feb,
  year      =  2013,
  url       = "https://iopscience.iop.org/article/10.1088/1742-5468/2013/02/P02003/meta",
  doi       = "10.1088/1742-5468/2013/02/P02003",
  issn      = "1742-5468"
}

@ARTICLE{Zou2024-ms,
  title     = "Introduction to dynamical mean-field theory of randomly connected
               neural networks with bidirectionally correlated couplings",
  author    = "Zou, Wenxuan and Huang, Haiping",
  journal   = "SciPost Phys. Lect. Notes",
  publisher = "Stichting SciPost",
  number    =  79,
  month     =  "20~" # feb,
  year      =  2024,
  url       = "https://scipost.org/SciPostPhysLectNotes.79/pdf",
  doi       = "10.21468/scipostphyslectnotes.79",
  issn      = "2590-1990"
}

@BOOK{Helias2020-kg,
  title     = "Statistical field theory for neural networks",
  author    = "Helias, Moritz and Dahmen, David",
  publisher = "Springer Nature",
  address   = "Cham, Switzerland",
  edition   =  1,
  series    = "Lecture notes in physics",
  month     =  "21~" # aug,
  year      =  2020,
  url       = "https://link.springer.com/book/10.1007/978-3-030-46444-8",
  doi       = "10.1007/978-3-030-46444-8",
  isbn      = "9783030464431,9783030464448",
  issn      = "0075-8450,1616-6361"
}

@InProceedings{Phuong2021-qh,
  title = 	 {Towards Understanding Knowledge Distillation},
  author =       {Phuong, Mary and Lampert, Christoph},
  booktitle = 	 {Proceedings of the 36th International Conference on Machine Learning},
  pages = 	 {5142--5151},
  year = 	 {2019},
  editor = 	 {Chaudhuri, Kamalika and Salakhutdinov, Ruslan},
  volume = 	 {97},
  series = 	 {Proceedings of Machine Learning Research},
  month = 	 {09--15 Jun},
  publisher =    {PMLR},
  url = 	 {https://proceedings.mlr.press/v97/phuong19a.html}
}

@InProceedings{Das2023-zx,
  title = 	 {Understanding Self-Distillation in the Presence of Label Noise},
  author =       {Das, Rudrajit and Sanghavi, Sujay},
  booktitle = 	 {Proceedings of the 40th International Conference on Machine Learning},
  pages = 	 {7102--7140},
  year = 	 {2023},
  editor = 	 {Krause, Andreas and Brunskill, Emma and Cho, Kyunghyun and Engelhardt, Barbara and Sabato, Sivan and Scarlett, Jonathan},
  volume = 	 {202},
  series = 	 {Proceedings of Machine Learning Research},
  month = 	 {23--29 Jul},
  publisher =    {PMLR},
  url = 	 {https://proceedings.mlr.press/v202/das23d.html}
}

@InProceedings{pmlr-v267-das25b,
  title = 	 {Retraining with Predicted Hard Labels Provably Increases Model Accuracy},
  author =       {Das, Rudrajit and Dhillon, Inderjit S and Epasto, Alessandro and Javanmard, Adel and Mao, Jieming and Mirrokni, Vahab and Sanghavi, Sujay and Zhong, Peilin},
  booktitle = 	 {Proceedings of the 42nd International Conference on Machine Learning},
  pages = 	 {12509--12538},
  year = 	 {2025},
  editor = 	 {Singh, Aarti and Fazel, Maryam and Hsu, Daniel and Lacoste-Julien, Simon and Berkenkamp, Felix and Maharaj, Tegan and Wagstaff, Kiri and Zhu, Jerry},
  volume = 	 {267},
  series = 	 {Proceedings of Machine Learning Research},
  month = 	 {13--19 Jul},
  publisher =    {PMLR},
  url = 	 {https://proceedings.mlr.press/v267/das25b.html}
}

@InProceedings{pmlr-v283-mandal25a,
  title = 	 {A Theoretical Analysis of Soft-Label vs Hard-Label Training in Neural Networks},
  author =       {Mandal, Saptarshi and Lin, Xiaojun and Srikant, Rayadurgam},
  booktitle = 	 {Proceedings of the 7th Annual Learning for Dynamics \&amp; Control Conference},
  pages = 	 {1078--1089},
  year = 	 {2025},
  editor = 	 {Ozay, Necmiye and Balzano, Laura and Panagou, Dimitra and Abate, Alessandro},
  volume = 	 {283},
  series = 	 {Proceedings of Machine Learning Research},
  month = 	 {04--06 Jun},
  publisher =    {PMLR},
  url = 	 {https://proceedings.mlr.press/v283/mandal25a.html}
}

@ARTICLE{Ji2020-ws,
  title     = "Knowledge distillation in wide neural networks: Risk bound, data
               efficiency and imperfect teacher",
  author    = "Ji, Guangda and Zhu, Zhanxing",
  editor    = "Larochelle, H and Ranzato, M and Hadsell, R and Balcan, M F and
               Lin, H",
  journal   = "Neural Inf Process Syst",
  publisher = "Curran Associates, Inc.",
  volume    = "abs/2010.10090",
  pages     = "20823--20833",
  month     =  "20~" # oct,
  year      =  2020,
  url       = "https://proceedings.neurips.cc/paper_files/paper/2020/hash/ef0d3930a7b6c95bd2b32ed45989c61f-Abstract.html",
  eprint    = "2010.10090"
}

@ARTICLE{Chen2017-ty,
  title     = "Learning efficient object detection models with knowledge
               distillation",
  author    = "Chen, Guobin and Choi, Wongun and Yu, Xiang and Han, T and
               Chandraker, Manmohan",
  journal   = "Neural Inf Process Syst",
  publisher = "proceedings.neurips.cc",
  volume    =  30,
  pages     = "742--751",
  month     =  "4~" # dec,
  year      =  2017,
  url       = "https://proceedings.neurips.cc/paper/2017/hash/e1e32e235eee1f970470a3a6658dfdd5-Abstract.html"
}

@inproceedings{calderon-etal-2023-systematic,
    title = "A Systematic Study of Knowledge Distillation for Natural Language Generation with Pseudo-Target Training",
    author = "Calderon, Nitay  and
      Mukherjee, Subhabrata  and
      Reichart, Roi  and
      Kantor, Amir",
    editor = "Rogers, Anna  and
      Boyd-Graber, Jordan  and
      Okazaki, Naoaki",
    booktitle = "Proceedings of the 61st Annual Meeting of the Association for Computational Linguistics (Volume 1: Long Papers)",
    month = jul,
    year = "2023",
    address = "Toronto, Canada",
    publisher = "Association for Computational Linguistics",
    url = "https://aclanthology.org/2023.acl-long.818/",
    doi = "10.18653/v1/2023.acl-long.818",
    pages = "14632--14659"
}

@INPROCEEDINGS{Liu2018-ib,
  title     = "Multi-label image classification via knowledge distillation from
               weakly-supervised detection",
  author    = "Liu, Yongcheng and Sheng, Lu and Shao, Jing and Yan, Junjie and
               Xiang, Shiming and Pan, Chunhong",
  booktitle = "Proceedings of the 26th ACM international conference on
               Multimedia",
  publisher = "ACM",
  address   = "New York, NY, USA",
  month     =  "15~" # oct,
  year      =  2018,
  url       = "https://dl.acm.org/doi/abs/10.1145/3240508.3240567?casa_token=nydQnM9Cf98AAAAA:deU8hY0sjEIDgwfhDu5GTQ3A50TR0f4aPiwz-ImXrtJHolNoj9WsCxmYxwdMATFGGrvlSUjj0TKYUw",
  doi       = "10.1145/3240508.3240567",
  isbn      =  9781450356657
}

@INCOLLECTION{Xu2020-am,
  title     = "Feature normalized knowledge distillation for image
               classification",
  author    = "Xu, Kunran and Rui, Lai and Li, Yishi and Gu, Lin",
  booktitle = "Lecture Notes in Computer Science",
  publisher = "Springer International Publishing",
  address   = "Cham",
  pages     = "664--680",
  series    = "Lecture notes in computer science",
  year      =  2020,
  url       = "https://link.springer.com/chapter/10.1007/978-3-030-58595-2_40",
  doi       = "10.1007/978-3-030-58595-2\_40",
  isbn      = "9783030585945,9783030585952",
  issn      = "0302-9743,1611-3349"
}

@inproceedings{
Gu2023-pl,
title={Mini{LLM}: Knowledge Distillation of Large Language Models},
author={Yuxian Gu and Li Dong and Furu Wei and Minlie Huang},
booktitle={The Twelfth International Conference on Learning Representations},
year={2024},
url={https://openreview.net/forum?id=5h0qf7IBZZ}
}

@inproceedings{hahn-choi-2019-self,
    title = "Self-Knowledge Distillation in Natural Language Processing",
    author = "Hahn, Sangchul  and
      Choi, Heeyoul",
    editor = "Mitkov, Ruslan  and
      Angelova, Galia",
    booktitle = "Proceedings of the International Conference on Recent Advances in Natural Language Processing (RANLP 2019)",
    month = sep,
    year = "2019",
    address = "Varna, Bulgaria",
    publisher = "INCOMA Ltd.",
    url = "https://aclanthology.org/R19-1050/",
    doi = "10.26615/978-954-452-056-4_050",
    pages = "423--430"
}

@inproceedings{clark-etal-2019-bam,
    title = "{BAM}! Born-Again Multi-Task Networks for Natural Language Understanding",
    author = "Clark, Kevin  and
      Luong, Minh-Thang  and
      Khandelwal, Urvashi  and
      Manning, Christopher D.  and
      Le, Quoc V.",
    editor = "Korhonen, Anna  and
      Traum, David  and
      M{\`a}rquez, Llu{\'i}s",
    booktitle = "Proceedings of the 57th Annual Meeting of the Association for Computational Linguistics",
    month = jul,
    year = "2019",
    address = "Florence, Italy",
    publisher = "Association for Computational Linguistics",
    url = "https://aclanthology.org/P19-1595/",
    doi = "10.18653/v1/P19-1595",
    pages = "5931--5937"
}

@ARTICLE{Zhang2020-bh,
  title     = "Self-Distillation as Instance-Specific Label Smoothing",
  author    = "Zhang, Zhilu and Sabuncu, M",
  editor    = "Larochelle, H and Ranzato, M and Hadsell, R and Balcan, M F and
               Lin, H",
  journal   = "Neural Inf Process Syst",
  publisher = "Curran Associates, Inc.",
  volume    = "abs/2006.05065",
  pages     = "2184--2195",
  month     =  "9~" # jun,
  year      =  2020,
  url       = "https://proceedings.neurips.cc/paper_files/paper/2020/hash/1731592aca5fb4d789c4119c65c10b4b-Abstract.html",
  eprint    = "2006.05065"
}

@inproceedings{
Pareek2024-qf,
title={Understanding the Gains from Repeated Self-Distillation},
author={Divyansh Pareek and Simon Shaolei Du and Sewoong Oh},
booktitle={The Thirty-eighth Annual Conference on Neural Information Processing Systems},
year={2024},
url={https://openreview.net/forum?id=gMqaKJCOCB}
}

@InProceedings{Gerace2020-bo,
  title = 	 {Generalisation error in learning with random features and the hidden manifold model},
  author =       {Gerace, Federica and Loureiro, Bruno and Krzakala, Florent and Mezard, Marc and Zdeborova, Lenka},
  booktitle = 	 {Proceedings of the 37th International Conference on Machine Learning},
  pages = 	 {3452--3462},
  year = 	 {2020},
  editor = 	 {III, Hal Daumé and Singh, Aarti},
  volume = 	 {119},
  series = 	 {Proceedings of Machine Learning Research},
  month = 	 {13--18 Jul},
  publisher =    {PMLR},
  url = 	 {https://proceedings.mlr.press/v119/gerace20a.html}
}

@ARTICLE{Loureiro2022-wp, 
  title     = "Learning curves of generic features maps for realistic datasets
               with a teacher-student model",
  author    = "Loureiro, Bruno and Gerbelot, C\'{e}dric and Cui, Hugo and Goldt,
               Sebastian and Krzakala, Florent and M\'{e}zard, Marc and
               Zdeborov\'{a}, Lenka",
  journal   = "J. Stat. Mech.",
  publisher = "IOP Publishing",
  volume    =  2022,
  number    =  11,
  pages     =  114001,
  month     =  "1~" # nov,
  year      =  2022,
  url       = "https://iopscience.iop.org/article/10.1088/1742-5468/ac9825/meta",
  doi       = "10.1088/1742-5468/ac9825",
  issn      = "1742-5468"
}

@INPROCEEDINGS{Zhang2019-lq,
  title     = "Be your own teacher: Improve the performance of convolutional
               neural networks via self distillation",
  author    = "Zhang, Linfeng and Song, Jiebo and Gao, Anni and Chen, Jingwei
               and Bao, Chenglong and Ma, Kaisheng",
  booktitle = "2019 IEEE/CVF International Conference on Computer Vision (ICCV)",
  publisher = "IEEE",
  pages     = "3713--3722",
  month     =  oct,
  year      =  2019,
  url       = "http://openaccess.thecvf.com/content_ICCV_2019/html/Zhang_Be_Your_Own_Teacher_Improve_the_Performance_of_Convolutional_Neural_ICCV_2019_paper.html",
  doi       = "10.1109/iccv.2019.00381",
  isbn      =  9781728148038
}

@INPROCEEDINGS{Lelarge2019-gx,
  title     = "Asymptotic Bayes risk for Gaussian mixture in a semi-supervised
               setting",
  author    = "Lelarge, Marc and Miolane, Leo",
  booktitle = "2019 IEEE 8th International Workshop on Computational Advances in
               Multi-Sensor Adaptive Processing (CAMSAP)",
  publisher = "IEEE",
  pages     = "639--643",
  month     =  dec,
  year      =  2019,
  url       = "https://ieeexplore.ieee.org/document/9022623?denied=",
  doi       = "10.1109/camsap45676.2019.9022623",
  isbn      = "9781728155494,9781728155487"
}

@inproceedings{
Ma2021-ga,
title={Undistillable: Making A Nasty Teacher That {\{}CANNOT{\}} teach students},
author={Haoyu Ma and Tianlong Chen and Ting-Kuei Hu and Chenyu You and Xiaohui Xie and Zhangyang Wang},
booktitle={International Conference on Learning Representations},
year={2021},
url={https://openreview.net/forum?id=0zvfm-nZqQs}
}

@ARTICLE{11014106,
  author={Yilmaz, Eda and Keles, Hacer Yalim},
  journal={IEEE Access}, 
  title={Adversarial Sparse Teacher: Defense Against Distillation-Based Model Stealing Attacks Using Adversarial Examples}, 
  year={2025},
  volume={13},
  number={},
  pages={92074-92085},
  doi={10.1109/ACCESS.2025.3573105}}

@INCOLLECTION{Gordon1988-mn,
  title     = "On Milman's inequality and random subspaces which escape through
               a mesh in $\mathbb{R}$ n",
  author    = "Gordon, Y",
  booktitle = "Lecture Notes in Mathematics",
  publisher = "Springer Berlin Heidelberg",
  address   = "Berlin, Heidelberg",
  pages     = "84--106",
  series    = "Lecture notes in mathematics",
  year      =  1988,
  url       = "https://link.springer.com/chapter/10.1007/BFb0081737",
  doi       = "10.1007/bfb0081737",
  isbn      = "9783540193531,9783540392354",
  issn      = "0075-8434,1617-9692"
}

@ARTICLE{Donoho2009-lr,
  title     = "Message-passing algorithms for compressed sensing",
  author    = "Donoho, David L and Maleki, Arian and Montanari, Andrea",
  journal   = "Proc. Natl. Acad. Sci. U. S. A.",
  publisher = "Proceedings of the National Academy of Sciences",
  volume    =  106,
  number    =  45,
  pages     = "18914--18919",
  month     =  "10~" # nov,
  year      =  2009,
  url       = "https://www.pnas.org/doi/abs/10.1073/pnas.0909892106",
  doi       = "10.1073/pnas.0909892106",
  pmc       = "PMC2767368",
  pmid      =  19858495,
  issn      = "0027-8424,1091-6490"
}

@ARTICLE{Thrampoulidis2015-ex,
  title     = "Regularized linear regression: A precise analysis of the
               estimation error",
  author    = "Thrampoulidis, Christos and Oymak, Samet and Hassibi, B",
  editor    = "Gr{\"{u}}nwald, Peter and Hazan, Elad and Kale, Satyen",
  journal   = "Conf Learn Theory",
  publisher = "PMLR",
  volume    =  40,
  pages     = "1683--1709",
  month     =  "26~" # jun,
  year      =  2015,
  url       = "https://proceedings.mlr.press/v40/Thrampoulidis15.html"
}

@ARTICLE{Krzakala2007-nu,
  title     = "Landscape analysis of constraint satisfaction problems",
  author    = "Krzakala, Florent and Kurchan, Jorge",
  journal   = "Phys. Rev. E Stat. Nonlin. Soft Matter Phys.",
  publisher = "American Physical Society (APS)",
  volume    =  76,
  number    = "2 Pt 1",
  pages     =  021122,
  month     =  aug,
  year      =  2007,
  url       = "https://journals.aps.org/pre/abstract/10.1103/PhysRevE.76.021122",
  doi       = "10.1103/PhysRevE.76.021122",
  pmid      =  17930021,
  issn      = "1539-3755,1550-2376"
}

@ARTICLE{Okajima2024-ba,
doi = {10.1088/1742-5468/add1ce},
url = {https://doi.org/10.1088/1742-5468/add1ce},
year = {2025},
month = {jun},
publisher = {IOP Publishing},
volume = {2025},
number = {5},
pages = {053301},
author = {Okajima, Koki and Takahashi, Takashi},
title = {Asymptotic dynamics of alternating minimization for bilinear regression},
journal = {Journal of Statistical Mechanics: Theory and Experiment}
}

@article{Dobriban2018high,
 ISSN = {00905364, 21688966},
 URL = {https://www.jstor.org/stable/26542784},
 author = {Edgar Dobriban and Stefan Wager},
 journal = {The Annals of Statistics},
 number = {1},
 pages = {247--279},
 publisher = {Institute of Mathematical Statistics},
 title = {HIGH-DIMENSIONAL ASYMPTOTICS OF PREDICTION: RIDGE REGRESSION AND CLASSIFICATION},
 urldate = {2024-02-07},
 volume = {46},
 year = {2018}
}

@InProceedings{pesce2023are,
  title = 	 {Are Gaussian Data All You Need? {T}he Extents and Limits of Universality in High-Dimensional Generalized Linear Estimation},
  author =       {Pesce, Luca and Krzakala, Florent and Loureiro, Bruno and Stephan, Ludovic},
  booktitle = 	 {Proceedings of the 40th International Conference on Machine Learning},
  pages = 	 {27680--27708},
  year = 	 {2023},
  editor = 	 {Krause, Andreas and Brunskill, Emma and Cho, Kyunghyun and Engelhardt, Barbara and Sabato, Sivan and Scarlett, Jonathan},
  volume = 	 {202},
  series = 	 {Proceedings of Machine Learning Research},
  month = 	 {23--29 Jul},
  publisher =    {PMLR},
  url = 	 {https://proceedings.mlr.press/v202/pesce23a.html}
}

@inproceedings{loureiro2021learning,
 author = {Loureiro, Bruno and Sicuro, Gabriele and Gerbelot, Cedric and Pacco, Alessandro and Krzakala, Florent and Zdeborov\'{a}, Lenka},
 booktitle = {Advances in Neural Information Processing Systems},
 editor = {M. Ranzato and A. Beygelzimer and Y. Dauphin and P.S. Liang and J. Wortman Vaughan},
 pages = {10144--10157},
 publisher = {Curran Associates, Inc.},
 title = {Learning Gaussian Mixtures with Generalized Linear Models: Precise Asymptotics in High-dimensions},
 url = {https://proceedings.neurips.cc/paper_files/paper/2021/file/543e83748234f7cbab21aa0ade66565f-Paper.pdf},
 volume = {34},
 year = {2021}
}

@inproceedings{
mannelli2022unfair,
title={Bias-inducing geometries: exactly solvable data model with fairness implications},
author={Stefano Sarao Mannelli and Federica Gerace and Negar Rostamzadeh and Luca Saglietti},
booktitle={ICML 2024 Workshop on Geometry-grounded Representation Learning and Generative Modeling},
year={2024},
url={https://openreview.net/forum?id=oupizzpMpY}
}

@book{doi:10.1142/13341,
author = {Charbonneau, Patrick and Marinari, Enzo and Mézard, Marc and Parisi, Giorgio and Ricci-Tersenghi, Federico and Sicuro, Gabriele and Zamponi, Francesco},
title = {Spin Glass Theory and Far Beyond},
publisher = {WORLD SCIENTIFIC},
year = {2023},
doi = {10.1142/13341},
address = {},
edition   = {},
URL = {https://www.worldscientific.com/doi/abs/10.1142/13341},
eprint = {https://www.worldscientific.com/doi/pdf/10.1142/13341}
}

@book{doi:10.1142/0271,
author = {Mezard, M and Parisi, G and Virasoro, M},
title = {Spin Glass Theory and Beyond},
publisher = {WORLD SCIENTIFIC},
year = {1986},
doi = {10.1142/0271},
address = {},
edition   = {},
URL = {https://www.worldscientific.com/doi/abs/10.1142/0271},
eprint = {https://www.worldscientific.com/doi/pdf/10.1142/0271}
}

@software{vaibhav_kumar_dixit_2023_7738525,
	author = {Vaibhav Kumar Dixit and Christopher Rackauckas},
	month = mar,
	publisher = {Zenodo},
	title = {Optimization.jl: A Unified Optimization Package},
	version = {v3.12.1},
	doi = {10.5281/zenodo.7738525},
  	url = {https://doi.org/10.5281/zenodo.7738525},
	year = 2023}

@ARTICLE{Hospedales2022-ap,
  title     = "Meta-learning in neural networks: A survey",
  author    = "Hospedales, Timothy and Antoniou, Antreas and Micaelli, Paul and
               Storkey, Amos",
  journal   = "IEEE Trans. Pattern Anal. Mach. Intell.",
  publisher = "Institute of Electrical and Electronics Engineers (IEEE)",
  volume    =  44,
  number    =  9,
  pages     = "5149--5169",
  month     =  sep,
  year      =  2022,
  url       = "http://dx.doi.org/10.1109/TPAMI.2021.3079209",
  doi       = "10.1109/TPAMI.2021.3079209",
  pmid      =  33974543,
  issn      = "0162-8828,1939-3539"
}

@ARTICLE{Blitzer2006-vu,
  title     = "Domain adaptation with structural correspondence learning",
  author    = "Blitzer, John and McDonald, Ryan T and Pereira, Fernando C",
  editor    = "Jurafsky, Dan and Gaussier, Eric",
  journal   = "Empir Method Nat Lang Process",
  publisher = "Association for Computational Linguistics",
  pages     = "120--128",
  month     =  "22~" # jul,
  year      =  2006,
  url       = "http://dx.doi.org/10.3115/1610075.1610094",
  doi       = "10.3115/1610075.1610094"
}

@INPROCEEDINGS{Bengio2009-gy,
  title     = "Curriculum learning",
  author    = "Bengio, Yoshua and Louradour, J\'{e}r\^{o}me and Collobert, Ronan
               and Weston, Jason",
  booktitle = "Proceedings of the 26th Annual International Conference on
               Machine Learning",
  publisher = "ACM",
  address   = "New York, NY, USA",
  month     =  "14~" # jun,
  year      =  2009,
  url       = "http://dx.doi.org/10.1145/1553374.1553380",
  doi       = "10.1145/1553374.1553380",
  isbn      =  9781605585161
}

@InProceedings{pmlr-v235-cui24d,
  title = 	 {Asymptotics of feature learning in two-layer networks after one gradient-step},
  author =       {Cui, Hugo and Pesce, Luca and Dandi, Yatin and Krzakala, Florent and Lu, Yue and Zdeborova, Lenka and Loureiro, Bruno},
  booktitle = 	 {Proceedings of the 41st International Conference on Machine Learning},
  pages = 	 {9662--9695},
  year = 	 {2024},
  editor = 	 {Salakhutdinov, Ruslan and Kolter, Zico and Heller, Katherine and Weller, Adrian and Oliver, Nuria and Scarlett, Jonathan and Berkenkamp, Felix},
  volume = 	 {235},
  series = 	 {Proceedings of Machine Learning Research},
  month = 	 {21--27 Jul},
  publisher =    {PMLR},
  url = 	 {https://proceedings.mlr.press/v235/cui24d.html}
}

@software{torchvision2016,
    title        = {TorchVision: PyTorch's Computer Vision library},
    author       = {TorchVision maintainers and contributors},
    year         = 2016,
    journal      = {GitHub repository},
    publisher    = {GitHub},
    howpublished = {\url{https://github.com/pytorch/vision}}
}

@article{Krizhevsky2009-ac,
  title={Learning multiple layers of features from tiny images},
  author={Krizhevsky, Alex and Hinton, Geoffrey and others},
  year={2009},
  publisher={Toronto, ON, Canada},
  url      = "https://www.cs.toronto.edu/~kriz/learning-features-2009-TR.pdf",
}

@article{takahashi2024a,
  title={A replica analysis of under-bagging},
  author={Takashi Takahashi},
  journal={Transactions on Machine Learning Research},
  year={2024},
  issn={2835-8856},
  url={https://openreview.net/forum?id=7HIOUZAoq5},
  archivePrefix={arXiv},
  eprint={2404.09779}
}

@inproceedings{yang-etal-2024-self,
    title = "Self-Distillation Bridges Distribution Gap in Language Model Fine-Tuning",
    author = "Yang, Zhaorui  and
      Pang, Tianyu  and
      Feng, Haozhe  and
      Wang, Han  and
      Chen, Wei  and
      Zhu, Minfeng  and
      Liu, Qian",
    editor = "Ku, Lun-Wei  and
      Martins, Andre  and
      Srikumar, Vivek",
    booktitle = "Proceedings of the 62nd Annual Meeting of the Association for Computational Linguistics (Volume 1: Long Papers)",
    month = aug,
    year = "2024",
    address = "Bangkok, Thailand",
    publisher = "Association for Computational Linguistics",
    url = "https://aclanthology.org/2024.acl-long.58/",
    doi = "10.18653/v1/2024.acl-long.58",
    pages = "1028--1043"
}

@misc{chen2025visiontransformersselfdistilledregisters,
      title={Vision Transformers with Self-Distilled Registers}, 
      author={Yinjie Chen and Zipeng Yan and Chong Zhou and Bo Dai and Andrew F. Luo},
      year={2025},
      eprint={2505.21501},
      archivePrefix={arXiv},
      primaryClass={cs.CV},
      url={https://arxiv.org/abs/2505.21501}, 
}

@INPROCEEDINGS{10619631,
  author={Akhtiamov, Danil and Ghane, Reza and Hassibi, Babak},
  booktitle={2024 IEEE International Symposium on Information Theory (ISIT)}, 
  title={Regularized Linear Regression for Binary Classification}, 
  year={2024},
  volume={},
  number={},
  pages={202-207},
  doi={10.1109/ISIT57864.2024.10619631}}

@inproceedings{kini2020analytic,
  title={Analytic study of double descent in binary classification: The impact of loss},
  author={Kini, Ganesh Ramachandra and Thrampoulidis, Christos},
  booktitle={2020 IEEE International Symposium on Information Theory (ISIT)},
  pages={2527--2532},
  year={2020},
  organization={IEEE}
}

@article{deng2022model,
  title={A model of double descent for high-dimensional binary linear classification},
  author={Deng, Zeyu and Kammoun, Abla and Thrampoulidis, Christos},
  journal={Information and Inference: A Journal of the IMA},
  volume={11},
  number={2},
  pages={435--495},
  year={2022},
  publisher={Oxford University Press}
}

@article{liu2024unifying,
  title={Unifying AMP Algorithms for Rotationally-Invariant Models},
  author={Liu, Songbin and Ma, Junjie},
  journal={arXiv preprint arXiv:2412.01574},
  year={2024}
}

\newpage
\appendix

\section{Further remarks on related works} \label{further_related_works}

\textbf{Replica method for machine learning problems.}
As machine learning models and datasets grow increasingly complex, traditional mathematical approaches often fall short in providing rigorous analytical solutions. 
This complexity gap has led to a rising demand for alternative theoretical tools that can offer insights into model behavior and performance, even when rigorous mathematical solutions are out of reach.

In this context, the replica method, originally developed in statistical physics, has emerged as a powerful analytical technique for machine learning problems.
While not yet mathematically rigorous in all aspects, this method has been widely applied to various models, from simple perceptrons~\cite{Gardner1988-lo} to modern non-i.i.d. datasets~\cite{Gerace2020-bo, Loureiro2022-wp}, with some of its predictions later rigorously proven~\cite{Barbier2019-wh}. 
The replica method offers unique advantages, such as the ability to compute exact generalization errors rather than bounds or necessary conditions. 
This precision enables explicit optimization of hyperparameters in multi-stage SD, providing deeper insights into model behavior and performance.

\textbf{Relationship between multi-stage replica method and DMFT.}
Traditional DMFT~\cite{Helias2020-kg} is primarily used for analyzing learning dynamics. 
This approach is effective when the system's state at time $t$ can be expressed explicitly using the state at time $t-1$, allowing for direct averaging over data. 
However, in more complex scenarios like SD, conventional DMFT techniques face challenges. In our model, the transition from one state to the next is not explicitly defined but is implicitly determined through an optimization process. 
Specifically, the previous state ($\hat{\vb{w}}^{t-1}, \hat{B}^{t-1}$) influences the output labels $y^t$ (Eq.~\eqref{eq:soft_label}), which then feed into the optimization problem minimizing the loss function (Eq.~\eqref{eq:loss_t}) to determine the new state ($\hat{\vb{w}}^{t}, \hat{B}^{t}$).
This implicit dependency, mediated by an optimization step, makes the dynamics more complex than those typically handled by traditional DMFT approaches. 
To overcome these challenges and extend DMFT's applicability to such complex scenarios, we employ the replica method, which allows us to analyze these implicit optimization-based state transitions effectively.

\newpage

\section{Further remarks on model and notation} \label{appendix:notation}
In this appendix, we compile and summarize all notation used throughout the paper, provide detailed commentary on the model's structure, and include illustrative figures to aid the reader's understanding.
\subsection{Notation}
\label{append:notation}
Table~\ref{tab:notation} lists each symbol and its definition.
\begin{table}[H]
  \centering
  \label{tab:notation}
  \caption{Summary of Notations}
  \begin{tabular}{lll}
    \toprule
    \textbf{Category} & \textbf{Symbol} & \textbf{Definition} \\
    \midrule
    \textit{Data Generation} 
      & $N$ & Dimension of input features \\
      & $M$ & Number of training samples \\
      & $\alpha = M/N$ & Data-to-dimension ratio, $\mathcal{O}(1)$ \\
      & $\vb{x}_\mu$ & Input vector ($\mu=1,\dots,M$) \\
      & $\rho$ & Class prior ($\rho\in(0,0.5]$) \\
      & $y_\mu^{\mathrm{true}}$ & True label, $\{0,1\}$ with $p(y^{\mathrm{true}})=\rho^{y}(1-\rho)^{1-y}$ \\
      & $y_\mu$ & Noisy observed label; $\Pr[y_\mu\neq y_\mu^{\mathrm{true}}]=\theta$ \\
      & $\theta$ & Label noise rate ($\theta\in (0, 0.5]$) \\
      & $\vb{v}$ & Gaussian mean vector, set $(1,1,\dots,1)^\top$ w.l.o.g. \\
      & $\Delta$ & Feature noise variance in \eqref{eq:gaussian_mixture}, $\mathcal{O}(1)$ \\
      & $D_{\rm tr}$ & Training set $\{(\vb{x}_\mu,y_\mu)\}_{\mu=1}^M$ \\
    \midrule
    \textit{Distillation Process}
      & $t$ & Stage index ($0$ = base model) \\
      & $(\vb{w}^t,B^t)$ & Weights and bias at stage $t$ (minimize \eqref{eq:loss_t}) \\
      & $y_\mu^t$ & Target label at stage $t$: $y_\mu^0=y_\mu$, for $t>0$ see \eqref{eq:soft_label} \\
      & $\lambda^t$ & $L_2$-regularization strength in \eqref{eq:loss_t} \\
      & $\beta^t$ & Inverse temperature for soft pseudo-label hardness in \eqref{eq:soft_label} \\
    \midrule
    \textit{Loss \& Prediction}
      & $\ell(y,\hat y)$ & Loss function: \{cross-entropy, MSE\} \\
      & $\sigma(x)$ & Activation: sigmoid $1/(1+e^{-x})$ or linear $(x+1)/2$ \\
      & $Y(\vb{w}^t,B^t;\vb{x})$ & Activation output, see Eq.~\eqref{eq:prediction} \\
      & $\mathcal{L}_t(\vb{w}^t,B^t)$ & Stage-$t$ objective, see Eq.~\eqref{eq:loss_t} \\
    \midrule
    \textit{Performance Metrics}
      & $\mathcal{E}^t$ & Generalization error after stage $t$, def.\ in Sec.~\ref{sec:problem_setup} \\
      & $\mathcal{E}^{*0}$ & Optimal $0$-SD error (minimized over $\lambda^0$) \\
      & $\mathcal{E}^{*t}$ & Optimal $t$-SD error (minimized over $\lambda^{0..t},\beta^{1..t}$) \\
      & $\mathcal{E}^{*t}_{\mathrm{Hard}}$ & Optimal $t$-SD error ($\beta^{1..t} \to \infty$ and minimized over $\lambda^{0..t}$) \\
    \midrule
    \textit{Asymptotic Quantities}
      & $Q^{st}$ & Weight–weight overlap, limit of $\hat w^s\cdot\hat w^t / N$ \\
      & $m^t$ & Weight–signal overlap, $\hat w^t\cdot v / N$ \\
      & $b^t$ & Rescaled bias, $\hat B^t/\sqrt{Q^{tt}}$ \\
    \bottomrule
  \end{tabular}
\end{table}

\subsection{Illustration of the model}
\label{append:illustration}
\begin{figure}[htb]
    \begin{center}
    \centerline{\includegraphics[width=1.0\columnwidth]{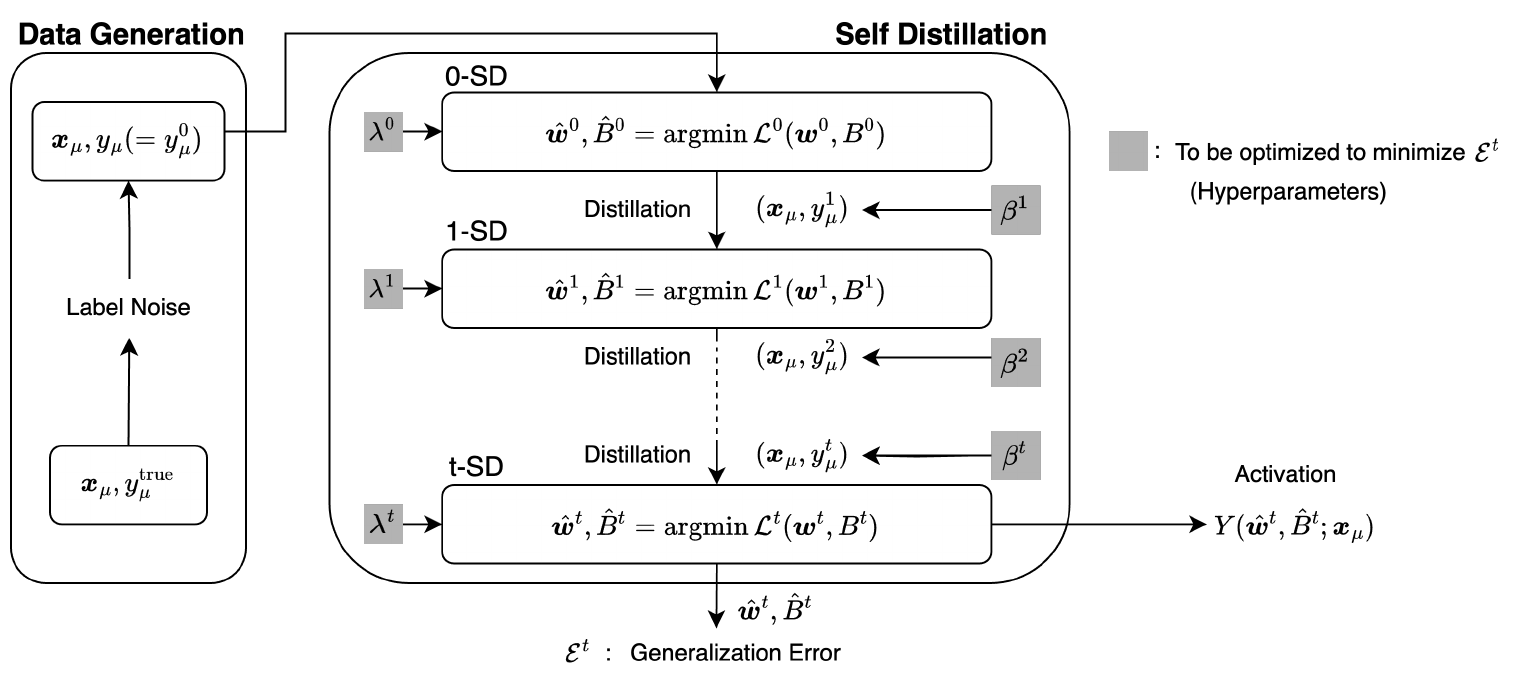}}
    \vskip -0.1in
    \caption{A schematic diagram of the $t$-SD model.}
    \label{fig:sketch}
    \end{center}
    \vskip -0.2in
\end{figure}

We present a graphical illustration of the model. 
Figure~\ref{fig:sketch} depicts the overall architecture, highlighting its key components and the flow of information.

\subsection{Motivation for the choice of activation function}
\label{append:activation_function}
We chose the activation function $\sigma(x) = (x+1)/2$ instead of the simpler $\sigma(x) = x$ because it ensures that the decision boundary remains unchanged when adjusting the temperature parameter.
This choice isolates the effects of soft labels, avoiding confounding influences from shifting decision boundaries.

\section{Replica calculation} \label{append:replica}
This appendix outlines the procedure for deriving Result~\ref{thm:replica} using the replica method. The process can be summarized in the following steps:

\begin{itemize}
    \item Present the complete statements of the results, including Results~\ref{thm:replica} (Subsection~\ref{subsection:full_theorem}).
    \item Introduce the average of the $p$-th moment of $\hat{\vb{w}}^1, \cdots, \hat{\vb{w}}^t$, denoted as $\mathcal{F}_{\vb{p}}^t$, which characterizes macroscopic quantities such as generalization error (Subsection~\ref{subsec:what}).
    \item  Show that the joint probability distribution of $\hat{\vb{w}}^t$ and $\hat{B}^t$ coincides with the correlation function of a system obtained by duplicating the original system, referred to as the replica system (Subsection~\ref{subsec:onestage} for $t=0$ and Subsection~\ref{subsec:tstage} for $t>0$).
    \item Outline the procedure for evaluating $\mathcal{F}_{\vb{p}}^t$ by incorporating certain assumptions into the replica variables (Subsections~\ref{subsec:partion} and \ref{subsec:RS}).
    \item Derive equations to determine the parameters necessary for calculating $\mathcal{F}_{\vb{p}}^t$ (Subsections~\ref{subsec:saddle1} and \ref{subsec:saddle2}).
    \item Use these equations to derive Result~\ref{thm:full_theorem} (Subsection~\ref{subsec:proof}).
\end{itemize}

Finally, we present remarks on the rigorous proofs in Section~\ref{sec:remarks_on_rigorous}.

\subsection{Full Statement of the Result}
\label{subsection:full_theorem}
Here we give the complete statements of the results to be proved.

\begin{theorem}
    \label{thm:full_theorem}
    (\textbf{Statistics of the T-SD model})
    Under the proportional asymptotic limit ($N, M\to\infty$, constrained by $M/N \to \alpha \in(0,\infty)$), we have 
    \begin{align}
        &\hat{w}^0_i \overset{\mathrm{d}}{=} \frac{1}{\hat{Q}^{00} + \lambda^0}  \ab(\hat{m}^0 + \hat{\xi}^0) \label{eq:w1} \\
        &\hat{w}^T_i \overset{\mathrm{d}}{=} \frac{1}{\hat{Q}^{TT} + \lambda^T}  \ab(\hat{m}^T + \hat{\xi}^T - \sum_{s=0}^{T-1}\hat{Q}^{st}\hat{w}^s) \quad (T\geq 1) \label{eq:wt} \\
        & \frac{\hat{\vb{w}}^T\cdot\vb{x}_\mu}{\sqrt{N}} + \hat{B}^T \overset{\mathrm{d}}{=} h^T + z^T_* , \label{eq:preactivation} 
    \end{align}
    where the parameters satisfy the following equations:
    \begin{align}
    &\begin{cases}
        Q^{0t} &= \frac{\hat{m}^0 m^t + R^{0t}}{\lambda^0 + \hat{Q}^{00}}  \\ 
        Q^{st} &= \frac{\hat{m}^s m^t + R^{st} - \sum_{l=0}^{s-1} \hat{Q}^{ls}Q^{lt}}{\lambda^t + \hat{Q}^{tt}} \quad (t \geq s \geq 1) 
    \end{cases} \\
    &\begin{cases}
        R^{s0} &= \frac{\hat{\chi}^{0s}}{\hat{Q}^{00}+\lambda^0}  \\
        R^{st} &= \frac{\hat{\chi}^{st}-\sum_{l=0}^{t-1}\hat{Q}^{lt}R^{sl}}{\hat{Q}^{tt}+\lambda^t} \quad (t \geq 1) 
    \end{cases} \\
    &\begin{cases}
        m^0 &= \frac{\hat{m}^0}{\lambda^0 + \hat{Q}^{00}}  \\
        m^t &= \frac{ \hat{m}^t - \sum_{s=0}^{t-1}\hat{Q}^{st}m^s }{\lambda^t + \hat{Q}^{tt}} \quad (t\geq 1)
    \end{cases} \\
    &\begin{cases}
        \chi^{ss} &= \frac{1}{\lambda^s + \hat{Q}^{ss}}  \\
        \chi^{s,t+1} &= - \frac{\hat{Q}^{t,t+1}}{\lambda^{t+1} + \hat{Q}^{t+1, t+1}}  \chi^{st}  \quad (t\geq s) 
    \end{cases} \\
    &\begin{cases}
    \hat{Q}^{st} &= - \frac{\alpha}{\chi^{tt}} \mathbb{E}_{y,y^{\mathrm{true}}, \vb{\xi}}\ab[ \odv{z^t_*}{h^s} ]  \\
    \hat{m}^t &=  \frac{\alpha}{\Delta \chi^{tt}} \mathbb{E}_{y,y^{\mathrm{true}}, \vb{\xi}}\ab[(2y-1) z^t_*  ] \\
    \hat{\chi}^{st} &=  \frac{\alpha}{\Delta \chi^{ss}\chi^{tt}}  \mathbb{E}_{y,y^{\mathrm{true}}, \vb{\xi}}\ab[ z^s_* z^t_* ] \\
    \mathbb{E}_{y,y^{\mathrm{true}}, \vb{\xi}}\ab[z^t_*] &= 0 .
    \end{cases}
    \end{align}
    Here, $Q = (Q^{st})$ , $\chi = (\chi^{st})$ and $\hat{\chi} = (\hat{\chi}^{st})$ are symmetric matrices in $\mathbb{R}^{(T+1) \times (T+1)}$,
    and we introduced the following notations:
    \begin{align}
    &z^t_* = \argmin_{z^t} \ab[ \frac{(z^t)^2}{2\Delta\chi^{tt}} + \ell\ab(y^t, \sigma\ab(h^t + z^t))  ]  \\
    &\begin{cases}
        y^0 &= y \\
        y^t &=  \sigma\ab(\beta^{t-1} \ab(h^{t-1} + z^{t-1}_*))   \quad (t\geq 1) \\
    \end{cases} \\
    &\begin{cases}
    h^0 &\overset{\mathrm{d}}{=}  \xi^0 + (2y^{\mathrm{true}}-1)m^0 + b^0  \\
    h^t &\overset{\mathrm{d}}{=}  \xi^t + \sum_{s=1}^{t-1} \frac{\chi^{st}}{\chi_{ss}}  z^s_*  + (2y^{\mathrm{true}}-1)m^t + b^t  \quad (t \geq 1) ,\label{eq:ht}
    \end{cases}
    \end{align}
    where 
    $\vb{\xi}=(\xi^t)\in\R^{T+1} \sim \mathcal{N}(\vb{0}, \Delta Q)$, $\hat{\vb{\xi}}=(\hat{\xi}^t)\in\R^{T+1} \sim \mathcal{N}(\vb{0}, \hat{\chi})$, 
    and $y, y^{\rm true}\in\{0,1\}$ that are generated as $p(y, y^{\rm true})= p(y\mid y^{\rm true})p(y^{\rm true})$ with $p(y^{\rm true}) = \rho^{y^{\rm true}}(1-\rho)^{1-y^{\rm true}}, p(y\neq y^{\rm true}\mid y^{\rm true})=\theta$.
\end{theorem}

\subsection{What to calculate}\label{subsec:what}
Our primary interest lies in understanding how macroscopic quantities, such as the generalization error, behave under fluctuations in the training data. 
These macroscopic quantities can generally be expressed, excluding the bias, as functions of
\begin{align}
\frac{1}{N}\sum_i(\hat{w}_i^0)^{p^0}\cdots(\hat{w}_i^t)^{p^t},\label{eq:macro_val_terms}
\end{align}
where $p^0, \cdots, p^t \in \mathbb{N}\cup{0}$.
In the asymptotic limit $N\to\infty$, we expect that Eq.~\eqref{eq:macro_val_terms} converges with probability $1$ to their expected values 
\begin{align}
    \lim_{N\to\infty}   \frac{1}{N} \sum_i ({w}_i^{0})^{p^0} \cdots ({w}_i^{t})^{p^t} 
    =\lim_{N\to\infty}   \frac{1}{N}\mathbb{E}_\mathcal{D}\ab[ \sum_i ({w}_i^{0})^{p^0} \cdots ({w}_i^{t})^{p^t}], \quad \text{almost surely.} \label{eq:selfave}
\end{align}
This kind of concentration is called the self-averaging property \cite{doi:10.1142/0271} in the context of statistical mechanics. Although it is not obvious that the self-averaging property holds, this property has been proved in several convex optimization problems \cite{Thrampoulidis2015-ex}. 

Thus, the quantity we need to compute is
\begin{align}
\mathcal{F}_{\vb{p}}^t = \lim_{N\to\infty}\mathbb{E}_\mathcal{D}\ab[(\hat{w}_i^0)^{p^0}\cdots(\hat{w}_i^t)^{p^t}]\label{eq:macro_val_before},
\end{align}
where $\vb{p}=(p^0, \cdots, p^t)$.

Since the optimization problem for each time step $t$ in our model is convex, $\hat{\vb{w}}^t$ and $\hat{b}^t$ are deterministically determined as $(\hat{\vb{w}}^0, \hat{B}^0) \rightarrow(\hat{\vb{w}}^1, \hat{B}^1)\rightarrow \cdots \rightarrow(\hat{\vb{w}}^t, \hat{B}^t)$ given the data $\mathcal{D}=\ab\{ \vb{x}_\mu, y^{\text{true}}_\mu, y_\mu \}_\mu$.
However, to facilitate our analysis, we intentionally treat this deterministic process as a stochastic one. Specifically, we model the transition from $(\hat{\vb{w}}^{s-1}, \hat{B}^{s-1})$ to $(\hat{\vb{w}}^{s}, \hat{B}^{s})$ using the following distribution
\begin{align}
    \hat{\vb{w}}^{s}, \hat{B}^{s} \sim p(\vb{w}^{s}, B^{s} \mid \hat{\vb{w}}^{s-1}, \hat{B}^{s-1}, \mathcal{D}) &= \lim_{\gamma^s\to\infty} \frac{\exp\ab(-\gamma^s \mathcal{L}_s(\vb{w}^s, B^s ))}{Z^s} \quad (s=1, \cdots, t),  \label{eq:markov} 
\end{align}
where $\mathcal{L}_s$ is defined in Eq.~\eqref{eq:loss_t} and 
\begin{align}
    Z^s &= \int \d \vb{w}^s \d B^s\, \exp\ab(-\gamma^s \mathcal{L}_s(\vb{w}^s, B^s )) \quad (s=1, \cdots, t) \label{eq:partition}
\end{align}
is the marginal likelihood (partition function).
Observe that in the limit as $\gamma^s\to\infty$, the distribution concentrates on $\argmin_{\vb{w}^s, B^s} \mathcal{L}_s(\vb{w}^s, B^s)$. 
Similary, the case $t=0$ is defined as
\begin{align}
    \hat{\vb{w}}^{0}, \hat{B}^{0} \sim p(\vb{w}^{0}, B^{0} \mid \mathcal{D}) &= \lim_{\gamma^0\to\infty} \frac{\exp\ab(-\gamma^0 \mathcal{L}_0(\vb{w}^0, B^0 ))}{Z^0} \quad (s=1, \cdots, t),  \label{eq:markov_0} 
\end{align}
where 
\begin{align}
    Z^0 &= \int \d \vb{w}^0 \d B^0 \, \exp\ab(-\gamma^0 \mathcal{L}_0(\vb{w}^0, B^0 )) \label{eq:partition_0} .
\end{align}

Following the probabilistic interpretation of these dynamics, the quantity defined in Eq.~\eqref{eq:macro_val_before} can be reformulated as
\begin{align}
\mathcal{F}_{\vb{p}}^t = \mathbb{E}_{\mathcal{D}}\ab[\braket<\cdots\braket<\braket<w^t_i>^{p^t}_t \ab(w^{t-1}_i)^{p^{t-1}}>_{t-1}\cdots \ab(w^0_i)^{{p^0}}>_0 ], \label{eq:macro_val}
\end{align}
where $\braket<f(w^{s})>_s$ is expectation under the distribution $p(\vb{w}^s, B^s \mid \vb{w}^{s-1}, B^{s-1}, \mathcal{D})$ if $s>0$ and  $p(\vb{w}^0, B^0 \mid  \mathcal{D})$ if $s=0$.

In summary, our computational task is to calculate the data average of statistical quantities $\mathcal{F}_{\vb{p}}^t$ (Eq.~\eqref{eq:macro_val}) for a sequence of random variables $(\hat{\vb{w}}^0, \hat{B}^0) \rightarrow(\hat{\vb{w}}^1, \hat{B}^1)\rightarrow \cdots \rightarrow(\hat{\vb{w}}^t, \hat{B}^t)$ following a Markov process defined by Eqs.~\eqref{eq:markov} and \eqref{eq:markov_0}.

\subsection{One-stage replica method}
\label{subsec:onestage}
To grasp the outline of the calculation in the replica method, we first consider the case of $t=0$.
For simplicity of notation, in the calculations within Subsections~\ref{subsec:onestage} and \ref{subsec:tstage}, we treat $\vb{w}^t$ as one-dimensional, omitting the subscript $i$ from $w^t_i$. However, it is straightforward to extend this to the general $N$-dimensional case.

The data average of the $p$-th moment of the solution $\hat{w}^0$ (Eq.~\eqref{eq:markov_0}) is given by
\begin{align}
  \mathcal{F}^0_p = \mathrm{E}_\mathcal{D} \ab[(\hat{w}^0)^p] &= \mathrm{E}_\mathcal{D} \ab[ \int \d  w^0\d B^0 \, w^0 p(w^0, B^0 \mid \mathcal{D}) ]^p  \\
  &= \lim_{\gamma^0 \to \infty} \mathrm{E}_\mathcal{D} \ab[ \int \d  w^0\d B^0 \, w^0 \frac{\exp \ab(-\gamma^0 \mathcal{L}_0(w^0, B^0 ))}{Z^0}]^p.   \label{eq:moment_p}
\end{align}

Direct computation of this is challenging due to the presence of the marginal likelihood $Z^0$ in the denominator of Eq.~\eqref{eq:moment_p}. To circumvent this difficulty, we introduce the following identity that holds for any $p\in\mathbb{N}$:
\begin{align}
  \mathrm{E}_{\cal D}[(\hat{w}^0)^p]=\lim_{n^0\to 0}\lim_{\gamma^0 \to\infty}\frac{{\cal W}_p(n^0,\gamma^0)}{\Xi_p(n^0,\gamma^0)}
  \label{eq:replica_identity}
\end{align}
where
\begin{align}
  {\cal W}_p(n^0,\gamma^0)&=\mathrm{E}_{\cal D}\left[\left\{\int \d w^0\d B^0 \, w^0  \exp(-\gamma^0{\cal L}(w^0, B^0))\right\}^p(Z^0)^{n^0-p}\right]\label{eq:replica_identity_numerator},\\
  {\Xi}(n^0,\gamma^0)&=\mathrm{E}_{\cal D}[(Z^0)^{n^0}]\label{eq:replica_identity_denominator}.
\end{align}

For the expectation with respect to data, we resort to a calculation method known as replica method. First, we assume that $n^t\in\mathbb{N}$ and $n^t>p$,\footnote{
    This assumption is not consistent with taking the limit $n^0 \to 0$, as performed in Eq.~\eqref{eq:replica_identity}. Therefore, it remains necessary to verify whether the results obtained under the condition $n^0 > p$ can be correctly extrapolated to the regime where $n^0 \to 0$.
    While the mathematical validity of this analytic continuation has not yet been rigorously proven, no counterexamples to its validity have been identified so far, at least in cases where the optimization problem determining the parameters $\hat{w}^t$ is convex.
} and express \eqref{eq:replica_identity_numerator} and \eqref{eq:replica_identity_denominator} by using $n^t$-replicated variables $w_1^t,\ldots,w_{n^t}^t$ as
\begin{align}
  {\cal W}_p(n^0,\gamma^0)&=\mathrm{E}_{\cal D}\left[\int \d\bm{w}^0 \d \vb{B}^0\, w_1^0\cdots w_p^0\exp\ab(-\gamma^0\sum_{a_0=1}^{n_0}{\cal L}(w_{a_0}^0, B^0_{a_0}))\right]\\
  &=\int \d\bm{w}^0 \d \vb{B}^0\, w_1^0\cdots w_p^0 \, \mathrm{E}_{\cal D}\left[\exp\ab(-\gamma^0\sum_{a_0=1}^{n_0}{\cal L}(w_{a_0}^0, B^0_{a_0}))\right]\\
  {\Xi}(n^0,\gamma^0)&=\mathrm{E}_{\cal D}\left[\int \d\bm{w}^0 \d \vb{B}^0\, \exp\ab(-\gamma^0\sum_{a_0=1}^{n_0}{\cal L}(w_{a_0}^0, B^0_{a_0}))\right] \\
  &= \int \d\bm{w}^0 \d \vb{B}^0\,\mathrm{E}_{\cal D}\left[ \exp\ab(-\gamma^0\sum_{a_0=1}^{n_0}{\cal L}(w_{a_0}^0, B^0_{a_0}))\right] ,
\end{align}
where $\vb{w}^0$ and $\vb{B}^0$ are shorthands for $w^0_1, \cdots, w^0_{n^0}$ and $B^0_1, \cdots , B^0_{n^0}$, respectively.
These expression indicates that ${\cal W}_p(n^0,\gamma^0)\slash{\Xi}(n^0,\gamma^0)$ can be regarded as the expectation of the $p$-body correlation of replica variables obeying the joint distribution
\begin{align}
  p(\vb{w}^0 , \vb{B}^0)=  \lim_{\gamma^0 \to \infty} \frac{1}{\Xi(n^0,\gamma^0)}\mathrm{E}_{\cal D}\left[\exp\left(-\gamma^0\sum_{a_0=1}^{n^0}{\cal L}(w_{a_0}^0, B^0_{a_0})\right)\right]. \label{eq:replica_dist}
\end{align}
The primary challenge in our initial calculations was the necessity of averaging over the data, which significantly complicated the process. 
However, by employing the replica method, as shown in Eq.~\eqref{eq:replica_dist}, we effectively incorporate the data average into the probability distribution of the variables, thereby enabling us to derive their statistical properties.
The system that follows the probability distribution given by Eq.~\eqref{eq:replica_dist} is called a replica system.

\subsection{Extention to multi-stage replica method}
\label{subsec:tstage}
While the preceding analysis follows the conventional replica method prescription, the current scenario presents a unique challenge: the dependence of each estimator $\hat{w}^t$ on its predecessor $\hat{w}^{t-1}$ significantly increases the complexity of the problem.
To address this issue, we employ an innovative approach that involves the recursive application of the replica trick at each stage of the process. 

To illustrate this approach, let us consider how Eq.~\eqref{eq:replica_dist} evolves along step $t$ under our recursive methodology. 
First we define the following recursive function:
\begin{align}
    f_{t-1}(w^{t-1}) &=\braket<w^t>^{p^t}_t \\
    f_{s}(w^{s}) &= \braket<f_{s+1}(w^{s+1})\ab(w^{s+1})^{p^{s+1}}>_{s+1} \quad (0\leq s < t-1).
\end{align}
Then, what we want to calculate (Eq.~\eqref{eq:macro_val}) is expressed as
\begin{equation}
    \mathcal{F}^t_{\vb{p}} = \mathbb{E}_\mathcal{D} \braket<f_0(w^0) (w^0)^{p^0}>_0 .
\end{equation}
On the other hand, assuming that $n^0, \cdots, n^t\in\mathbb{N}$, $n^t>p$ and $n^{0}, \cdots, n^{t-1} > 1$, one can deduce
\begin{align}
    f_{t-1}(w^{t-1}) &= \braket<w^t>_t^{p^t} \\
    &= \ab\{ \int \d w^t\d B^t \, w^t \exp\ab(-\gamma^t \mathcal{L}_t(w^t, B^t)) \}^{p^t} \ab(Z^t)^{n^t - p^t}  \\
    &=  \int \d \vb{w}^t\d \vb{B}^t\,w^t_1 \cdots w^t_{p^t}   \exp\ab(-\gamma^t \sum_{a_t=1}^{n^t} \mathcal{L}_t(w^t_{a_t}, B^t_{a_t})) \label{eq:ft-1} \\
    f_{s-1}(w^{s-1}) &= \braket<f_s(w^s)(w^s)^{p^s}>_s \\
    &= \ab\{ \int \d w^s\d B^s \, f_s(w^s)(w^s)^{p^s}  \exp\ab(-\gamma^s \mathcal{L}_s(w^s, B^s)) \}^{p^s} \ab(Z^s)^{n^s - p^s}  \\
    &=  \int \d \vb{w}^s\d \vb{B}^s\, f_s(w^s_1)w^s_1 \cdots w^s_{p^s}   \exp\ab(-\gamma^s \sum_{a_s=1}^{n^s} \mathcal{L}_s(w^s_{a_s}, B^s_{a_s})) \quad (1\leq s < t) \label{eq:fs-1}\\
    \mathcal{F}^t_{\vb{p}} &= \braket<f_0(w^0)(w^0)^{p^0}>_0 \\
    &= \ab\{ \int \d w^0\d B^0 \, f_0(w^0)(w^0)^{p^0}  \exp\ab(-\gamma^0 \mathcal{L}_s(w^0, B^0)) \}^{p^0} \ab(Z^0)^{n^0 - p^0}  \\
    &=  \int \d \vb{w}^0\d \vb{B}^0\, f_0(w^0_1)w^0_1 \cdots w^0_{p^0}   \exp\ab(-\gamma^0\sum_{a_0=1}^{n^0} \mathcal{L}_0(w^0_{a_0}, B^0_{a_0})) \label{eq:Fp},
\end{align}
where $\vb{w}^t$ and $\vb{B}^t$ are shorthands for $w^t_1, \cdots, w^t_{n^t}$ and $B^t_1, \cdots , B^t_{n^t}$, respectively.
These equations demonstrate that when considering the parameters at time s while keeping the parameters at time $s-1$ fixed, the distribution of replica variables $\vb{w}^s$ and $\vb{B}^s$ at time $s$ depends on the first replica variables $w^{s-1}_1$ and $b^{s-1}_1$ at time $s-1$.
By recursively substituting Eqs.~\eqref{eq:ft-1} and \eqref{eq:fs-1} into \eqref{eq:Fp}, we have
\begin{align}
  \mathcal{F}_{\vb{p}}^t=\lim_{\substack{\gamma^0 \to \infty \\ n^0 \to 0}} \cdots \lim_{\substack{\gamma^t \to \infty \\ n^t \to 0}}   \frac{\mathcal{W}^t_{\vb{p}}(\vb{n}, \vb{\gamma})}{\Xi^t(\vb{n}, \vb{\gamma})}, 
  \label{eq:replica_identity_t}
\end{align}
where $\vb{n} = (n^0, \cdots, n^t)$, $\vb{\gamma} = (\gamma^0, \cdots, \gamma^t)$ and
\begin{align}
      {\cal W}_{\vb{p}}^t(\vb{n},\vb{\gamma})&=\int \d\bm{w}^0\cdots\d\vb{w}^t\d\vb{B}^0\cdots\d\vb{B}^t \, \ab((w_1^t\cdots w_{p^t}^t) \times(w^{t-1}_1)^{p^{t-1}}\times \cdots\times(w^0_1)^{p^0})\\
  &\qquad\qquad\qquad \times \mathbb{E}_{\cal D}\left[\exp\ab(-\sum_{s=1}^t \gamma^s\sum_{a_s=1}^{n_s}{\tilde{\mathcal{L}}}_s(w_{a^s}^s, B_{a^s}^s)) \right]\\
  {\Xi}^t(\vb{n},\vb{\gamma})&=\mathbb{E}_\mathcal{D} \ab[(Z^t)^{n^{t}}\times (Z^{t-1})^{n^{t-1}}\times  \cdots \times(Z^0)^{n^0} ] \\
  &=\mathbb{E}_{\cal D}\left[\int \d\bm{w}^0\cdots\d\vb{w}^t\d\vb{B}^0\cdots\d\vb{B}^t\, \exp\ab(-\sum_{s=1}^t \gamma^s\sum_{a_s=1}^{n_s}{\tilde{\mathcal{L}}}_s(w_{a^s}^s, B_{a^s}^s)) \right] \\
  &=\int \d\bm{w}^0\cdots\d\vb{w}^t\d\vb{B}^0\cdots\d\vb{B}^t\, \mathbb{E}_{\cal D}\left[\exp\ab(-\sum_{s=1}^t \gamma^s\sum_{a_s=1}^{n_s}{\tilde{\mathcal{L}}}_s(w_{a^s}^s, B_{a^s}^s)) \right], \label{eq:replica_partition_t}
\end{align}
where $\tilde{\mathcal{L}}(w^t_{a_t}, B^t_{a^t})$ represents the loss when the first replica at $t-1$ is used as a pseudo-label for training, i.e.,
\begin{align}
    \tilde{\mathcal{L}}(w^t_{a_t}, B^t_{a^t}) &= \sum_\mu \ell\ab(\tilde{y}^t_\mu, Y(w^t_{a_t}, B^t_{a_t}; x_\mu)) + \frac{\lambda^t}{2}\|w^t_{a_t} \|^2 \\
    \tilde{y}^t_\mu &= \sigma\ab(\beta^t\ab(\frac{\hat{w}^{t-1}_1\cdot x_\mu + \hat{B}^{t-1}_1} {\sqrt{N}})) \quad (t>0) \\
    \tilde{y}^0_\mu &= y_\mu  .
\end{align}

These expression indicates that ${\cal W}^t_{\vb{p}}(\vb{n},\vb{\gamma})\slash{\Xi}^t(\vb{n},\vb{\gamma})$ can be regarded as the expectation of the $(p^0+\cdots+p^t)$-body correlation of replica variables obeying the joint distribution
\begin{align}
  p(\vb{w}^0, \cdots, \vb{w}^t, \vb{B}^0, \cdots, \vb{B}^t)  = \lim_{\substack{\gamma^0 \cdots \gamma^{t} \to \infty}} \frac{1}{\Xi^t(\vb{n},\vb{\gamma})}\mathrm{E}_{\cal D}\left[\exp\left(-\sum_{s=1}^t \gamma^s\sum_{a_s=1}^{n_s}{\tilde{\mathcal{L}}}_s(w_{a^s}^s, B^s_{a^s})\right) \right]. \label{eq:replica_dist_t}
\end{align}
Ultimately, our problem reduces to calculating the statistical properties of the replica variables $\vb{w}^0, \cdots, \vb{w}^t$  that follow the distribution given by Eq.~\eqref{eq:replica_dist_t}.
Following the standard prescription for analysis in the asymptotic limit, it is crucial to investigate the behavior of the replica partition function Eq.~\eqref{eq:replica_partition_t} for this analysis.

\subsection{The calculation of the replica partition function}
\label{subsec:partion}
From this subsection, we recover the subscript of dimension $i$ in $w^t_{a_t, i}$.
Our next step is to calculate data average in the replica partition function (Eq.~\eqref{eq:replica_partition_t}).
To achieve this, we first calculate the partition function for a finite $n^1, \cdots, n^t$, and then consider the limit as $n^1, \cdots, n^t \to 0$.
 
First, we define the linearly transformed variable $\vb{u}^t = (u_1^t, \cdots, u_{a_t}^t, \cdots, u_{n^t}^t)^T$ as
\begin{equation}
  \vb{u}^t = \sqrt{\frac{\Delta}{N}} \sum_i \vb{w}_i^t z_i,
\end{equation}
where $z_i$ is standard normal random variavle defined in Eq.~\eqref{eq:gaussian_mixture} and $\vb{w}^t_i = (w^t_{1,i}, \cdots, w^t_{a_t,i}, \cdots, w^t_{n^t,i})^T$.
Then, $\vb{u} = \ab((\vb{u}^1)^T, \cdots, (\vb{u}^T)^T)$ also follows a Gaussian distribution, with the mean and covariance given by
\begin{equation}
  \mathbb{E}_\mathcal{D} \ab[ {u}^t_{a_t} ] = {0}, \quad \mathbb{E}_\mathcal{D} \ab[ {u}^s_{a_s} {u}^t_{c_t} ] = \Delta Q^{st}_{a_sc_t}, \label{eq:u_dist_beforeRS} 
\end{equation}
where $Q^{st}_{a_sc_t}$ is defined as
\begin{equation}
  Q^{st}_{a_sc_t} = \frac{1}{N} \vb{w}^s_{a_s} \cdot \vb{w}^t_{c_t}.
\end{equation}

Then, the partition function of the replica distribution Eq.~\eqref{eq:replica_partition_t} can be expressed as
  
\begin{align}
  \Xi^T(\vb{n}, \vb{\gamma}) &= 
  \int \vb{w}^0\cdots\vb{w}^T\int\d\vb{B}^0\cdots\vb{B}^T \, \exp\ab(-\sum_t\frac{\lambda^t}{2}\sum_{a_t}\|\vb{w}^t_{a_t}\|^2) \notag\\
  & \qquad \times \prod_{\mu=1}^M \mathbb{E}_{\vb{x}_\mu, y_\mu, y^{\text{true}}_\mu} \exp\ab[-\sum_{t}\sum_{a_t} \gamma^t \ell\ab(\tilde{y}^t_\mu, \sigma\ab(\frac{\vb{w}^t_{a_t}\cdot\vb{x}_\mu}{\sqrt{N}} + B^t_{a_t} ))] \\
  &=\int \d\vb{Q}\d\vb{m}\int \vb{w}^0\cdots\vb{w}^T\int\d\vb{B}^0\cdots\vb{B}^T\, \ab[\prod_{st}\prod_{a_s ,c_t} \delta\ab(NQ_{a_s c_t}^{st} - \vb{w}^s_{a_s}\vb{w}^t_{c_t})] \notag\\
  &\qquad \times \ab[\prod_{t}\prod_{a_t}\delta\ab(Nm^t_{a_t} - \vb{w}^t_{a_t}\cdot\vb{v})] \notag\\
  &\qquad \times\prod_{i=1}^N\exp\ab(-\sum_t\frac{\lambda^t}{2}\sum_{a_t}|w^t_{a_t, i}|^2) \notag\\
  &\qquad \times \prod_{\mu=1}^M \mathbb{E}_{\vb{x}_\mu, y_\mu, y^{\text{true}}_\mu} \exp\ab[-\sum_{t}\sum_{a_t} \gamma^t \ell\ab(\tilde{y}^t_\mu, \sigma\ab(\frac{\vb{w}^t_{a_t}\cdot\vb{x}_\mu}{\sqrt{N}} + B^t_{a_t} ))]  \label{eq:partition2},
\end{align}
where $\int\d\vb{Q}$ is an integration over $\{Q_{a_s c_t}^{st}\}_{1\leq s \leq t\leq T, 1\leq a_s\leq n^s, 1\leq c_t \leq n^t}$ and $\int \d\vb{m}$ is 
 an integration over $\{m^t_{a_t}\}_{1\leq t\leq T, 1\leq a_t \leq n^t}$.
 Using the following integral representations of the Dirac delta function\footnote{The integrations in Eqs.~\eqref{eq:delta1} and \eqref{eq:delta2} are performed along the imaginary axis.}:
 \begin{align}
     \delta\ab(NQ^{st}_{a_s c_t}-\vb{w}^s_{a_s}\cdot\vb{w}^t_{c_t}) &= \int \d\hat{{{Q}}}^{st}_{a_sc_t} \,\exp\ab(-\frac{\hat{Q}^{st}_{a_sc_t}}{2}\ab(NQ^{st}_{a_s c_t}-\vb{w}^s_{a_s}\cdot\vb{w}^t_{c_t})) \label{eq:delta1}\\
     \delta(Nm^t_{t_a} - \vb{v}\cdot\vb{w}^t_{a_t}) &= \int \d\hat{m}^{t}_{a_t}\, \exp\ab(-\hat{m}^t_{a_t}(Nm^t_{a_t}-\vb{v}\cdot\vb{w}^t_{a_t})).\label{eq:delta2}
 \end{align}
 One can find that Eq.~\eqref{eq:partition2} can be separated into three terms: (1) an interaction term $G_I$, which shows the interaction between order parameters (parameters without hat) and conjugate parameters (parameters with hat); (2) an entropic term $G_S$, which scales as $N$; and (3) an energy term $G_E$, which scales as $M$. 
 Based on these observations, calculating each component of the partition function yields the following equation:
\begin{align}
    \Xi^T(\vb{n}, \vb{\gamma}) 
 = \int \d\vb{Q}\d\vb{m}\d\hat{\vb{Q}}\d\hat{\vb{m}}\d\vb{B}^0\cdots\d\vb{B}^T\, (G_I)^N \ab(G_S)^N \ab(G_E)^{M} \label{eq:Zn}
\end{align}
where 
\begin{equation}
  G_I = \exp\ab[ -\ab( \frac{1}{2}\sum_{st}\sum_{a_sc_t}\hat{Q}_{a_sc_t}^{st}{Q}_{a_sc_t}^{st} + \sum_t\sum_{a_t} \hat{m}_{a_t} m_{a_t}  ) ]  \label{eq:GI}
\end{equation}

\begin{equation}
  G_S = \int \d \vb{w}^0\cdots\vb{w}^T \exp\ab[ -\sum_t \sum_a \frac{\gamma^t}{2}\lambda^t \ab(w_a^t)^2 + \sum_t\sum_a \hat{m}_{a_t}^t w_{a_t}^t  + \frac{1}{2}\sum_{st}\sum_{a_sc_t}\hat{Q}_{a_sc_t}^{st} w_{a_s}^s w_{c_t}^t  ]  \label{eq:GS}
\end{equation}

\begin{align}
  G_E &= \mathrm{E}_{y^{\text{true}}, y}\mathrm{E}_{\vb{u}}  \, \exp\ab[-\gamma^0 \sum_{a_0} \ell \ab(y, \sigma\ab((2y^\text{true}-1)m_{a_0}^0 + u_{a_0}^0 + B_{a_0}^0)) ] \notag \\
  &\quad \times \prod_{t=1}^T  \exp\ab[-\gamma^t \sum_{a_t} \ell \ab(\sigma\ab((2y^\text{true}-1)m_1^{t-1} + u_1^{t-1} + B_1^{t-1})   , \sigma\ab((2y^\text{true}-1)m_{a_t}^t + u_{a_t}^t + B_{a_t}^t)) ] . \label{eq:GE}
\end{align}

In the asimptotic limit ($N, M\to\infty, \alpha = M/N = \mathcal{O}(1)$),  Eq.~\eqref{eq:Zn} can evaluated using the saddle point method. 
Using this technique, the partition function (Eq.~\eqref{eq:Zn}) is evaluated as
\begin{align}
    \Xi^T(\vb{n},\vb{\gamma}) &= \exp\ab[N \max\ab[\Psi(\vb{Q},\vb{m},\hat{\vb{Q}},\hat{\vb{m}}, \vb{B}^0,\cdots, \vb{B}^T)]]\\
    &=\exp\ab[N \ab[\Psi(\vb{Q}^*,\vb{m}^*,\hat{\vb{Q}}^*,\hat{\vb{m}}^*, {\vb{B}^0}^*,\cdots, {\vb{B}^T}^*)]],  \label{eq:saddle_eval}
\end{align}
where 
\begin{equation}
    \Psi(\vb{Q},\vb{m},\hat{\vb{Q}},\hat{\vb{m}}, \vb{B}^0,\cdots, \vb{B}^T) = \log G_I + \log G_S + \alpha \log G_E.
\end{equation}
and 
\begin{equation}
    \vb{Q}^*,\vb{m}^*,\hat{\vb{Q}}^*,\hat{\vb{m}}^*, {\vb{B}^0}^*,\cdots, {\vb{B}^T}^* = \argmax_{\vb{Q},\vb{m},\hat{\vb{Q}},\hat{\vb{m}}, \vb{B}^0,\cdots, \vb{B}^T} \Psi(\vb{Q},\vb{m},\hat{\vb{Q}},\hat{\vb{m}}, \vb{B}^0,\cdots, \vb{B}^T).
\end{equation}
The saddle point equations to minimize $\Psi(\vb{Q},\vb{m},\hat{\vb{Q}},\hat{\vb{m}}, \vb{B}^0,\cdots, \vb{B}^T)$ are given by
\begin{equation}
  Q^{st}_{a_sc_t} = \frac{1}{G_S} \pdv{G_S}{\hat{Q}^{st}_{a_sc_t}}, \quad  m^t_{a_t} = \frac{1}{G_S} \pdv{G_S}{\hat{m}^t_{a_t}}, \quad \hat{Q}^{st}_{a_sc_t} = \alpha \frac{1}{G_E} \pdv{G_E}{Q^{st}_{a_sc_t}}, \quad  \hat{m}^t_{a_t} = \alpha \frac{1}{G_E} \pdv{G_E}{m^t_{a_t}}, \label{eq:saddle_naive}
\end{equation}
from the saddle point conditions of $\hat{Q}^{st}_{a_sc_t}$, $\hat{m}^s_{a_t}$, ${Q}^{st}_{a_sc_t}$ and ${m}^s_{a_t}$, respectively, and
\begin{align}
    \frac{1}{G_E}\pdv{G_E}{B^t_{a_t}} = 0,
\end{align}
from the saddle point condition of $B^t_{a_t}$.

In the limit where $n^1, \cdots n^T\to 0$, we have $G_E \to 1$ and $G_S \to 1$. Therefore, Eq.~\eqref{eq:saddle_naive} specifically become
\begin{align}
  Q_{a_sc_t}^{a_sc_t} &= \mathbb{E}_{\vb{w}} \ab[w^s_{a_s} w^t_{c_t}] \label{eq:saddle_Q}\\
  m_{a_t}^t &= \mathbb{E}_{\vb{w}} \ab[w^t_{a_t}] \\
  \hat{Q}_{a_sc_t}^{st} &= 2\alpha \pdv{}{Q_{a_sc_t}^{st}} \mathbb{E}_{\vb{u},y, y^\text{true}} \exp\ab[-\sum_{t=0}^T \gamma^t \sum_{a_t} \ell\ab(v^{t-1}_1, v^t_{a_t})] \label{eq:saddle_m}\\
  &= \Delta\alpha \mathbb{E}_{\vb{u}, y, y^\text{true}}\pdv[]{}{u^s_{a_s} u^t_{c_t}}  \exp\ab[-\sum_{t=0}^T \gamma^t \sum_{a_t} \ell\ab(v^{t-1}_1, v^t_{a_t})] \\
  \hat{m}_{a_t}^t &= \alpha \pdv{}{m^t_{a_t}} \mathbb{E}_{\vb{u},y, y^\text{true}} \exp\ab[-\sum_{t=0}^T \gamma^t \sum_{a_t} \ell\ab(v^{t-1}_1, v^t_{a_t})] \\
  &= \alpha \mathbb{E}_{\vb{u},y, y^\text{true}}\pdv{}{u^t_{a_t}}  \exp\ab[-\sum_{t=0}^T \gamma^t \sum_{a_t} \ell\ab(v^{t-1}_1, v^t_{a_t})] ,
\end{align}
where we assumed the random variables $\ab\{w^t_{a_t}\}_{t,a^t}$ in Eqs.~\eqref{eq:saddle_Q} and \eqref{eq:saddle_m} follow the following distribution:
\begin{equation}
  p(\vb{w}) \propto \exp\ab[\sum_t\sum_{a_t}\frac{\gamma^t}{2}\lambda^t \ab(w^t_{a_t})^2 + \sum_t\sum_{a_t} \hat{m}^t_{a_t} w^t_{a_t}  + \frac{1}{2}\sum_{st}\sum_{a_sc_t}\hat{Q}^{st}_{a_sc_t} w^s_{a_s} w^t_{c_t}] , \label{eq:w_dist_beforeRS}
\end{equation}
and  
\begin{align}
    v^t_{a_t} &= \sigma\ab((2y^{\text{true}}-1)m^t_{a_t} + u^t_{a_t} + B^t_{a_t}) \quad (t\geq 0) \\
    v^{-1}_{1} &= y .
\end{align}

\subsection{RS assamption}
\label{subsec:RS}
From Eqs.~\eqref{eq:replica_dist_t} and \eqref{eq:saddle_eval}, the solutions of saddle point equations are related to the statistical properties of the replica variables, i.e.,
\begin{equation}
  m^t_{a_t} = \frac{1}{N}  \sum_{i}\mathbb{E}  \ab[{w}^t_{a_t, i}],  \quad  Q^{st}_{a_s c_t} = \frac{1}{N} \sum_{i}\mathbb{E}  \ab[{w}^{s}_{a_s,i} {w}_{c_t,i}^t]
\end{equation}
in $n^1,\cdots, n^T\to 0$ limit, with the expectation taken over the probability distribution defined by Eq.~\eqref{eq:replica_dist_t}.
From the fact that the $p$-body correlation functions between replicas correspond to the $p$-th moments $\mathcal{F}_{\vb{p}}$ in the original Markov process, we obtain 
\begin{align}
     &\mathcal{F}_{\vb{e}^{(t)}}=\mathbb{E}_{\mathcal{D}} \ab[\hat{w}_i^t] = m^t_{1}\\ 
     &\mathcal{F}_{\vb{e}^{(s, t)}}=\mathbb{E}_{\mathcal{D}} \ab[\hat{w}_i^s \hat{w}_i^t] = Q^{st}_{a_s c_t},\label{eq:mQ_rel}
\end{align}
where $\vb{e}^{(t)}$ and $\vb{e}^{(s,t)}$ are $T$-dimensional vectors defined as follows:  
\[
\vb{e}^{(t)} = (e_1, e_2, \dots, e_T), \quad \text{where } e_i = 
\begin{cases} 
1 & \text{if } i = t, \\ 
0 & \text{otherwise,}
\end{cases}
\]  
and  
\[
\vb{e}^{(s,t)} = (e_1, e_2, \dots, e_T), \quad \text{where } e_i = 
\begin{cases} 
1 & \text{if } i = s \text{ or } i = t, \\ 
0, & \text{otherwise,}
\end{cases}
\]
and $(a_s, a_t) = (1,2)$ if $s=t=T$, and $(a_s, a_t) = (1,1)$ otherwise. Equation~\eqref{eq:mQ_rel} holds for an arbitrary $i$ since the integrand of Eq.~\eqref{eq:Zn} can be written independently of $i$.

However, the replica parameters ($m^t_a, Q^{st}_a, \hat{m}^t_a$ and $\hat{Q}^{st}_{ab}$) are ill-defined in the limit as $n^1, \cdots, n^T \to 0$.
To further advance our calculations and obtain well-defined quantities, we invoke the replica symmetry (RS) assumption, which posits a symmetry under permutation between different replicas, i.e., 
\begin{align}
  m_a^t &= m^t \\
  Q_{a_tc_t}^{tt} &= Q^{tt} + \frac{\chi^{tt}}{\gamma^t} \delta_{a_tc_t} \\
  Q_{a_sc_t}^{st} &= Q^{st} + \frac{\chi^{st}}{\gamma^s} \delta_{a_s 1} \quad (s \neq t) \\
  \hat{m}_{a_t}^t &= \gamma^t \hat{m}^t \\
  \hat{Q}_{a_tc_t}^{tt} &= \ab(\gamma^t)^2 \hat{\chi}^{tt} - \gamma^t \hat{Q}^{tt} \delta_{a_sc_t} \\
  \hat{Q}_{a_sc_t}^{st} &= \gamma^s\gamma^t \hat{\chi}^{st} - \gamma^t \hat{Q}^{st} \delta_{{a_s}1} \quad (s \neq t) \\
  B^t_{a_t} &= b^t
\end{align}
where $\delta_{ab}$ is the Kronecker delta function. 
Using this parameterization, one can deduce 
\begin{align}
    m^t &= \mathbb{E}\ab[\hat{w}_i^t] \label{eq:1stat} \\
    Q^{st} &= \mathbb{E}\ab[\hat{w}^s_i \hat{w}^t_i], \label{eq:2stat}
\end{align}
for an arbitrary $i$.
Higher-order moments can be immediately determined to be zero from Eq.~\eqref{eq:Zn}, given that the distribution in Eq.~\eqref{eq:replica_dist_t} indicates that $\vb{w}^1, \cdots, \vb{w}^T$ follow a Gaussian distribution.
Although the RS assumption is not mathematically rigorous in general, it has been empirically validated in many practical scenarios, particularly in convex optimization problems. To date, there are no known examples where the RS assumption leads to incorrect predictions in convex settings.

\subsection{Saddle point equations: order parameters}
\label{subsec:saddle1}
Our next step is to derive the saddle-point equations for the order parameters ($m^t, Q^{st}$ and $\chi^{st}$) under the RS assamption from the ones for replica parameters ($\hat{m}^t_a, \hat{Q}^{st}_{ab}$ and $\hat{\chi}^{st}_{ab}$).
First, substitution of the RS assumption in eq.~\eqref{eq:w_dist_beforeRS} yields
\begin{align}
  p(\vb{w}) &\propto  \exp\left[-\sum_t \gamma_t \frac{\hat{Q}^{tt}+\lambda^t}{2} \sum_{a_t}\ab(w^t_{a_t})^2 + \sum_t\gamma^t \hat{m}^t \sum_{a_t} w_{a_t}^t + \sum_{(s<t)} \gamma^t\hat{Q}^{st}w^{a_s}_1\sum_{a_t} w^t_{a_t} \right. \notag \\
  & \hspace{15em}  \left. + \frac{1}{2}\sum_{st}\hat{\chi}^{st} \ab(\gamma^s\sum_{a_s} w_{a_s}^s)\ab(\gamma^t\sum_{a_t} w^t_{a_t})  \right]  \\
  &= \int \D\hat{\vb{\xi}}  \exp\left[-\sum_t \gamma_t \frac{\hat{Q}^{tt}+\lambda^t}{2} \sum_{a_t}\ab(w^t_a)^2 + \sum_t\gamma^t \hat{m}^t \sum_{a_t} w_a^t + \sum_{(s<t)} \gamma^t\hat{Q}^{st}w^s_1\sum_a w^t_{a_t} \right. \notag \\
  & \hspace{15em}  \left. + \sum_{st}\gamma_s\sum_{a_s} w^s_{a_s}\sqrt{\hat{\chi}}^{st}\xi^t \right]  \\
  &=   \int \D\hat{\xi}^0 \exp\ab[-\gamma^1\sum_{a_0}\ab(\frac{\hat{Q}^{00}+\lambda^0}{2}\ab(w_{a_0}^0)^2 - \ab(\hat{m}^1 + \sum_s \sqrt{\hat{\chi}}^{0s}\xi^s )w_{a_0}^0)] \notag\\
  &\times \prod_{t=1}^T \int \D\hat{\xi}^t \exp\ab[-\gamma^t\sum_{a_t}\ab(\frac{\hat{Q}^{tt}+\lambda^t}{2}\ab(w_{a_t}^t)^2 - \ab(\hat{m}^t + \sum_s \sqrt{\hat{\chi}}^{ts}\xi^s - \sum_{s=0}^{t-1}\hat{Q}^{st}w^s_1)w_{a_t}^t)] , \label{eq:w_dist}
\end{align}
where $\sqrt{\hat{\chi}}^{st}$ is the $s, t$ element of the cholesky decomposition of the matrix $\hat{\chi}$, and $\D\hat{\xi}^t$ is the normal Gaussian measure. 

Now we define following notations:
\begin{align}
  f^t(w_a^t; w^0_1, \dots, w^{t-1}_1) &=  \begin{cases}
    \frac{\hat{Q}^{00}+\lambda^0}{2}\ab(w_a^0)^2 - \ab(\hat{m}^0 + \sum_s \sqrt{\hat{\chi}}^{0s}\xi^s )w_a^0 \quad (t=0) \\
    \frac{\hat{Q}^{tt}+\lambda^t}{2}\ab(w_a^t)^2 - \ab(\hat{m}^t + \sum_s \sqrt{\hat{\chi}}^{ts}\xi^s - \sum_{s=1}^{t-1}\hat{Q}^{st}w^s_1)w_a^t \quad (t \geq 1) \\
  \end{cases} \\
  w^t_*(w^0_1,\cdots,w^{t-1}_1) &= \argmin_{w^t}f^t(w^t; w^0_1, \cdots, w^{t-1}_1) = \begin{cases}
    \frac{\hat{m}^0 + \sum_s \sqrt{\hat{\chi}}^{0s}\xi^s}{\lambda^1 + \hat{Q}^{00}} \quad (t=0) \\
    \frac{\hat{m}^t + \sum_s \sqrt{\hat{\chi}}^{ts}\xi^s - \sum_{s=0}^{t-1}\hat{Q}^{st}w^s_1}{\lambda^t + \hat{Q}^{tt}} \quad (t \geq 1) \\
  \end{cases} \\
  w^t_* &= w^t_*(w^0_*, w^1_*(w^0_*), \cdots, w^{t-1}_*(w^0_*, \cdots, w^{t-2}_*)) \\
  w^t_*(w^0, \cdots, w^{s}_1) &= w^t_*(w^0_1, \cdots, w^{s}_1, w^{s+1}_*(w^0, \cdots, w^s)) .
\end{align}
Then, the saddle-point equations of order parameters are simplified as (in the $n^1, \cdots, n^T \to 0, \gamma^1,\cdots,\gamma^T\to 0$ limit)
\begin{align}
  {m}^t &= \mathbb{E} \ab[w^t_a] \\
  &= \int\D\hat{\vb{\xi}}\, \prod_{s=0}^{t-1}\int\d\vb{w}^s\, \exp\ab(-\gamma^sf^s) \frac{\int\d w^t\, w^t\exp\ab(-\gamma^tf^t)}{\int\d w^t\, \exp\ab(-\gamma^tf^t)} \\
  &= \mathbb{E}_{\hat{\vb{\xi}}}\, \ab[w^t_*] 
\end{align}
\begin{align}
  {Q}^{st} &= \mathbb{E} \ab[w^s_{a_s} w^t_{c_t}] \\
  &= \int\D\hat{\vb{\xi}}\, \prod_{l=0}^{s-1}\int\d\vb{w}^l\, \exp\ab(-\gamma^lf^l) \frac{\int\d w^s\, w^s \exp\ab(-\gamma^sf^s)}{\int\d w^s\, \exp\ab(-\gamma^sf^s)}\notag \\
  & \hspace{5em} \times \prod_{l=s+1}^{t-1}\int\d\vb{w}^l\, \exp\ab(-\gamma^lf^l) \frac{\int\d w^t\, w^t\exp\ab(-\gamma^tf^t)}{\int\d w^t\, \exp\ab(-\gamma^tf^t)} \\
  &= \mathbb{E}_{\hat{\vb{\xi}}}\, \ab[w^s_* w^t_*] \quad (s < t)  
\end{align}
\begin{align}
  {Q}^{tt} &= \mathbb{E} \ab[w^t_{a_t} w^t_{c_t}] \\
  &= \int\D\hat{\vb{\xi}}\, \prod_{s=0}^{t-1}\int\d\vb{w}^s\, \exp\ab(-\gamma^sf^s) \ab(\frac{\int\d w^t\, w^t\exp\ab(-\gamma^tf^t)}{\int\d w^t\, \exp\ab(-\gamma^tf^t)})^2 \\ 
  &= \mathbb{E}_{\hat{\vb{\xi}}}\, \ab[(w^t_*)^2] 
\end{align}
\begin{align}
  {\chi}^{st} &= \gamma^s \mathbb{E} \ab[w^s_{a_s} w^t_{c_t} - w^s_1 w^t_{c_t}] \\
  &= \gamma^s \int\D\hat{\vb{\xi}}\, \prod_{l=0}^{s-1}\int\d\vb{w}^l\, \exp\ab(-\gamma^lf^l) \left[  \frac{\int\d w^s_1 \, w^s_1 \exp\ab(-\gamma^sf^s(w^s_1))}{\int\d w^s\, \exp\ab(-\gamma^sf^s(w^s))}w^t_*(w^1_1,\cdots,w^s_1) \right. \notag \\
  &\hspace{5em} \left. - \frac{\int\d w^s \, w^s\exp\ab(-\gamma^sf^s(w^s)) \int \d w^s_1\, w^t_*(w^1_1, \cdots,w^s_1)}{\ab(\int\d w^s\, \exp\ab(-\gamma^sf^s))^2} \right] \\
  &= \int\D\hat{\vb{\xi}}\, \prod_{l=0}^{s-1}\int\d\vb{w}^l\, \exp\ab(-\gamma^lf^l)  \odv{}{\hat{m}^s} \frac{\int \d w^s_1\, \exp\ab(-\gamma^sf^s(w^s_1)) w^t_*(w^1_1,\cdots,w^s_1)}{\int\d w^s\, \exp\ab(-\gamma^sf^s)} \\
  &= \mathbb{E}_{\hat{\vb{\xi}}}\, \ab[\odv{w^t_*}{\hat{m}^s}] \quad (s < t) 
\end{align}
\begin{align}
  {\chi}^{tt} &= \gamma^t \mathbb{E} \ab[\ab(w^t_{a_t})^2 - w^t_{a_t}w^t_{c_t}] \\
  &= \gamma^t \int\D\hat{\vb{\xi}} \, \prod_{s=0}^{t-1}\int\d\vb{w}^s\, \exp\ab(-\gamma^sf^s) \ab[\frac{\int\d w^t\, \ab(w^t)^2\exp\ab(-\gamma^tf^t)}{\int\d w^t\, \exp\ab(-\gamma^tf^t)} - \ab(\frac{\int\d w^t\, w^t\exp\ab(-\gamma^tf^t)}{\int\d w^t\, \exp\ab(-\gamma^tf^t)})^2] \\
  &= \int\D\hat{\vb{\xi}}\, \prod_{s=0}^{t-1}\int\d\vb{w}^s\, \exp\ab(-\gamma^sf^s) \pdv{}{\hat{m}^t} \frac{\int\d w^t\, w^t\exp\ab(-\gamma^tf^t)}{\int\d w^t\, \exp\ab(-\gamma^tf^t)} \\
  &= \mathbb{E}_{\hat{\vb{\xi}}}\, \ab[\pdv{w^t_*}{\hat{m}^t}] .
\end{align}

By introducing the helper variable $ R^{st} = \mathbb{E}_{\hat{\vb{\xi}}} \ab[\sqrt{\hat{\chi}}^{ts} \hat{\xi}^s \hat{w}^t_*]$ for simplicity, the explicit calculation of these equations yields
\begin{align}
  R^{st} &= \mathbb{E}\ab[\sqrt{\hat{\chi}}^{ts} \hat{\xi}^s \hat{w}^t_*] = 
  \begin{cases}
    \frac{1}{\hat{Q}^{00}+\lambda^{0}} \hat{\chi}^{s0} \quad (t=0) \\
    \frac{1}{\hat{Q}^{tt} + \lambda^{t}} \ab(\hat{\chi}^{st} - \sum_{l=0}^{t-1}\hat{Q}^{lt}R^{sl}) \quad (t \geq 1)
  \end{cases} \label{eq:saddle1}\\
  Q^{st} &= \mathbb{E}\ab[\hat{w}^s_* \hat{w}^t_*] =
  \begin{cases}
    \frac{1}{\hat{Q}^{00} + \lambda^0} \ab(\hat{m}^0 m^t + R^{0t}) \quad (s=0) \\
    \frac{1}{\hat{Q}^{ss} + \lambda^s} \ab(\hat{m}^sm^t + R^{st} - \sum_{l=0}^{s-1}\hat{Q}^{ls}Q^{lt}) \quad (s \geq 1) \\
  \end{cases}\label{eq:saddle2} \\
  m^t &= \mathbb{E}\ab[\hat{w}^t_*] = 
  \begin{cases}
    \frac{1}{\hat{Q}^{00} + \lambda^0}\hat{m}^0 \quad (t=0) \\
    \frac{1}{ \hat{Q}^{tt} + \lambda^t}\hat{m}^t \ab(\hat{m}^t - \sum_{s=0}^{t-1}\hat{Q}^{st}m^s) \quad (t \geq 1)
  \end{cases} \label{eq:saddle3}\\
  \chi^{st} &= \mathbb{E}\ab[\odv{w^t_*}{\hat{m}^s}] = 
  \begin{cases}
    \frac{1}{\hat{Q}^{tt} + \lambda^t}  \quad (s=t) \\
    -\frac{1}{\hat{Q}^{tt} + \lambda^{t}} \sum_{l=0}^s \hat{Q}^{lt} \chi^{sl} \quad (s < t)
  \end{cases} \label{eq:saddle4}.
\end{align}

\subsection{Saddle point equations: conjugate parameters}
\label{subsec:saddle2}

Similar to the previous section, we now derive the saddle-point equations for the conjugate parameters ($m^t, q^{st}$, $\chi^{st}$) based on the RS assumption.

The covariance matrix of the Gaussian variables $\vb{u}$ (Eq.~\eqref{eq:u_dist_beforeRS}) is rewritten as
\begin{align}
  \mathbb{E}_\mathcal{D} \ab[ u_{a_t}^t  u_{c_t}^t] &= \Delta \ab( Q^{tt} + \frac{\chi^{tt}}{\gamma^t} \delta_{a_tc_t} \ab)  \label{eq:u_cov} \\
  \mathbb{E}_\mathcal{D} \ab[ u_{a_s}^s u_{c_t}^t ] &= \Delta \ab( Q^{st} + \frac{\chi^{st}}{\gamma^s} \delta_{a_s1}  )  \quad (s < t) \label{eq:u_cov2} .
\end{align}
Under these conditions, we can introduce the random variable $\tilde{\vb{u}}$ with an equivalent distribution as follows:
\begin{equation}
  \tilde{u}^t_a = \sum_{r=0}^t A_{tr} \xi^r_0 + \sum_{r=0}^{t} \frac{\chi^{rt}}{\chi^{rr}} z^r_1 , 
\end{equation}
where $A_{st}$ are the cholesky decomposition of the covariance matrix of $\vb{u}$, i.e., $\sum_r A_{sr}A_{tr} = \Delta Q^{st}$, $z^t_a = \sqrt{\Delta \chi^{tt}/ \gamma^t} \xi^t_a$,
and $\xi^t_0, \xi^t_a \sim \mathcal{N}(0,1)$ are independent standard normal random variables. 

Following the same procedure as the previous section, taking the limit of $\gamma^t\to\infty$ in order, the expectation calculation is transformed into the solution of the optimization problem, 
and finally the following relationship is obtained:
\begin{align}
  \hat{Q}^{st} &= - \frac{\alpha}{\chi^{tt}} \mathbb{E}_{y,y^{\mathrm{true}}, \vb{\xi}}\ab[ \odv{z^t_*}{h^s} ] \label{eq:saddle5} \\
  \hat{m}^t &=  \frac{\alpha}{\Delta \chi^{tt}} \mathbb{E}_{y,y^{\mathrm{true}}, \vb{\xi}}\ab[(2y-1) z^t_*  ] \label{eq:saddle6}\\
  \hat{\chi}^{st} &=  \frac{\alpha}{\Delta \chi^{ss}\chi^{tt}}  \mathbb{E}_{y,y^{\mathrm{true}}, \vb{\xi}}\ab[ z^s_* z^t_* ] \label{eq:saddle7}\\
  &\mathbb{E}_{y,y^{\mathrm{true}}, \vb{\xi}}\ab[z^t_*] = 0 \label{eq:saddle8},
\end{align}
where $z^t_*$ is the solution of the optimization problem as follows:
\begin{align}
  z^0_* &= \argmin_{z^0} \ab[ \frac{(z^0)^2}{2\Delta\chi^{00}} + \ell\ab(y^0, h^0 + z^0)  ]  \label{eq:z0app}\\
  h^0 &=  A_{00} \xi^0_1 + (2y^\text{true}-1)m^0 + b^0  \\
  z^t_\ast &= \argmin_{z^t} \ab[ \frac{(z^t)^2}{2\Delta\chi^{tt}} + \ell\ab(\sigma\ab(\beta^t\ab(h^{t-1} + z^{t-1}_*)), \sigma\ab(h^t + z^t))  ]  \quad (1 \leq t \leq T) \label{eq:ztapp} \\
  h^t &= \sum_{s=0}^{t} A_{st} \xi^s_0 + \sum_{s=0}^{t-1} B_{st} z^s_*  + (2y^\text{true}-1)m^t + b^t  \quad (1 \leq t \leq T) .
\end{align}

\subsection{Derivation of Result~\ref{thm:full_theorem}}\label{subsec:proof}
To summarize the results obtained so far, we have derived the first- and second-order statistics of the estimator $\hat{w}^t_i$, which are determined by the constants $m^t$ and $Q^{st}$ (Eqs.~\eqref{eq:1stat} and \eqref{eq:2stat}). Furthermore, we have shown that these constants can be computed by solving the saddle-point equations defined in Eqs.~\eqref{eq:saddle1}--\eqref{eq:saddle4} and Eqs.~\eqref{eq:saddle5}--\eqref{eq:saddle8}. As a representation of the distribution of $\hat{w}^t_i$ that satisfies all the conditions for the integer moments, we can express it as  
\begin{align}
        &\hat{w}^0_i \overset{\mathrm{d}}{=} \frac{1}{\hat{Q}^{00} + \lambda^0}  \ab(\hat{m}^0 + \hat{\xi}^0) \\
        &\hat{w}^t_i \overset{\mathrm{d}}{=} \frac{1}{\hat{Q}^{tt} + \lambda^t}  \ab(\hat{m}^t + \hat{\xi}^t - \sum_{s=0}^{t-1}\hat{Q}^{st}\hat{w}^s) \quad (t\geq 1)
\end{align}

and this representation is unique up to equivalent forms.
Furthermore, by proceeding with similar calculations while taking into account the correlation with the data, Eqs.~\eqref{eq:z0app} and \eqref{eq:ztapp} yield
\begin{align}
    \frac{\hat{\vb{w}}^t\cdot\vb{x}_\mu}{\sqrt{N}} + \hat{B}^t \overset{\mathrm{d}}{=} h^t + z^t_*
\end{align}
for the pre-activation distribution.

\subsection{Remarks on Rigorous Proofs}
\label{sec:remarks_on_rigorous}
Beyond the replica method, a fully rigorous proof in our setting remains open. 
The main difficulty is the temporal correlation across multiple SD stages. 
AMP/state-evolution techniques \cite{Donoho2009-lr, liu2024unifying} have been applied successfully 
to handle such correlations, but basically only for first-order iterative algorithms, 
and thus do not directly extend to multi-stage SD. 
CGMT-based approaches \cite{Thrampoulidis2015-ex} could in principle be adapted, 
but they are typically used in on-line settings where new data are introduced at each stage. 
Developing tools that combine these techniques to rigorously capture the multi-stage dynamics 
would be an interesting direction for future work.

\subsection{Remarks on Numerical Calculations}

\paragraph{Iterative Method}
In numerical calculations of the saddle point equations Eqs.~\eqref{eq:saddle1}-\eqref{eq:saddle4} and Eqs.~\eqref{eq:saddle5}-\eqref{eq:saddle8} in replica method, the order parameters are usually obtained through iterative method. The procedure starts from an initial guess of the order parameters. These values are substituted into the left-hand side of the saddle-point equations, and the resulting right-hand side gives an updated estimate of the parameters. This process is repeated until the values converge within a chosen tolerance. In this way, the order parameters are determined as the fixed point of the equations.

\paragraph{Discrete Expectation}
When the theoretical formulation includes averages over discrete random variables such as the class label and the label-flip noise, each possible configuration is evaluated separately. In the present model, the class label 
$y$ takes values $0$ or $1$, and the label may be either flipped or unflipped. Hence, there are four possible combinations: $(y,y_{\text{true}}) = (0, 0), (0, 1), (1,0), (1,1)$. For a generic quantity $f(y,y_{\text{true}})$, the expectation appearing in the saddle-point equations is computed as
\begin{equation}
    \E_{y, y_{\text{true}}} \ab[f(y,y_{\text{true}})]=  \rho (1-\theta)f(0,0)+(1-\rho) \theta f(0, 1) + \rho (1-\theta) f(1, 0) + (1-\rho)(1-\theta)f(1,1).
\end{equation}
In the numerical implementation, the value of $f(y,y_{\text{true}})$ is computed and maintained for all four cases.

\paragraph{Sequential Update in Time}
The calculation also proceeds sequentially with respect to the time index $t$. The result at step $t$ depends only on the quantities determined at earlier steps from $0 \to t-1$. Therefore, the parameters can be updated step by step, starting from $t=0$. This recursive approach makes it straightforward to implement the computation and to monitor convergence at each stage.

\paragraph{Evaluation of Total Derivatives in Eq.~\eqref{eq:saddle5}}
We describe the implementation of evaluating the total derivatives appearing on the left-hand side of Eq.~\eqref{eq:saddle5} during the numerical calculation of the saddle point equations.

For notational simplicity, we define the partial derivatives of the loss function $\ell$ as
\begin{align}
  \ell'_t = \pdv{\ell(\sigma\ab(h^{t-1}+z^{t-1}_*), \sigma\ab(h^{t}+z^{t}))}{h^t}, \quad \bar{\ell}_t = \pdv{\ell(\sigma\ab(h^{t-1}+z^{t-1}_*), \sigma\ab(h^{t}+z^{t}))}{h^{t-1}}.
\end{align}
Introducing the function
\begin{align}
  F^t(z^t, z^{t-1}_*, h^t, h^{t-1}) = \frac{z^t}{\Delta \chi^{tt}} + \ell'_t,
\end{align}
we have the condition
\begin{align}
  F^t(z^t_*, z^{t-1}_*, h^t, h^{t-1}) = 0.
\end{align}
By employing the implicit function theorem, we obtain
\begin{align}
  \pdv{F^t}{z^t_*} \odv{z^t_*}{h^s} + \pdv{F^t}{h^{t-1}} \odv{h^{t-1}}{h^s} + \pdv{F^t}{h^t} \odv{h^t}{h^s} + \pdv{F^t}{z^{t-1}_*} \odv{z^{t-1}_*}{h^s} = 0.
\end{align}
since $z^t_*$ represents the solution of the optimization problem given by Eq.~\eqref{eq:ztapp}.
Solving for $\odv{z^t_*}{h^s}$, we obtain
\begin{align}
  \odv{z^t_*}{h^s} &= a^t \odv{h^t}{h^s} + b^t \odv{h^{t-1}}{h^s} + c^t \odv{z^{t-1}_*}{h^s},
\end{align}
where the coefficients $a^t, b^t, c^t$ are given by
\begin{align}
  a^t &= -\frac{\pdv{F^t}{h^t}}{\pdv{F^t}{z^t_*}} = - \frac{\ell''}{\ell'' + 1/\ab(\Delta \chi^{tt})},\\
  b^t &= -\frac{\pdv{F^t}{h^{t-1}}}{\pdv{F^t}{z^t_*}} = - \frac{\bar{\ell}'}{\ell'' + 1/\ab(\Delta \chi^{tt})}, \\
  c^t &= -\frac{\pdv{F^t}{z^{t-1}_*}}{\pdv{F^t}{z^t_*}} = - \frac{\bar{\ell}'}{\ell'' + 1/\ab(\Delta \chi^{tt})} = b^t.
\end{align}

Furthermore, using the following relations,
\begin{align}
  \odv{h^t}{h^s} &= \sum_{r < t} \frac{\chi^{rt}}{\chi^{rr}} \odv{z^r_*}{h^s}, \\
  \odv{z^{t}_*}{h^s} &= \sum_{r < t-1} \frac{\chi^{r,t-1}}{\chi^{r,t-1}} \odv{z^r_*}{h^s},
\end{align}
one can iteratively compute $\odv{z^t_*}{h^s}$ for a fixed $t$, as illustrated in Algorithm~\ref{algorithm_of_hatq}, which presents a single iteration step of the update procedure.
The full derivative is obtained by repeatedly applying this update until convergence.

\begin{algorithm}[H]
\label{algorithm_of_hatq}
  \caption{\textbf{UpdateQhatColumn$(t)$}: Self-consistent update of column $t$ (past columns are fixed)}
  \KwIn{
  Results up to time $t-1$:
  \begin{align}
    \ab\{\displaystyle z_*^{1:t-1},\ h^{1:t-1},\ G[s, r] = \tfrac{dz_*^r}{dh^s},\ H[s, r] = \tfrac{dh^r}{dh^s} \ (s \leq r \leq t-1)\}
  \end{align}
  }
  \KwOut{Results at time $t$:
  \begin{align}
    \ab\{\displaystyle G[s, t] = \tfrac{dz^t_*}{dh^s},\ H[s, t] = \tfrac{dh^t}{dh^s}\ (s \leq t),\ \hat{Q}[s, t]\ (s \leq t)\}
  \end{align}
  }
  \BlankLine
  
  \textbf{Step 1: Calculate $a^t, b^t, c^t$}\\
  \begin{align}
    a^t &\gets  - \frac{\ell''_{t,t}}{\ell''_{t,t} + 1/\ab(\Delta \chi^{t,t})}\\
    b^t &\gets - \frac{\bar{\ell'}_{t,t-1}}{\ell'_{t,t} + 1/\ab(\Delta \chi^{t,t})} \\
    c^t &\gets  b^t
  \end{align}

  \BlankLine

  \textbf{Step 2: Calculate $H[t, s]$}\\
  \For{$s \gets 1$ \KwTo $t-1$}{
    $H[s, t] \gets \sum_{r < t} \frac{\chi^{rt}}{\chi^{rr}} G[s, r]$
  }
  $H[t,t] \gets 1$

  \BlankLine

  \textbf{Step 3: Calculate $G[t, s]$}\\
  \For{$s \gets 1$ \KwTo $t-1$}{
    $G[s, t]\gets a^t H[s, t] + b^t H[s, t-1] + c^t G[s, t-1]$
  }
  $G[t,t] \gets a^t$

  \BlankLine

  \textbf{Step 5: Update $\hat Q[t, s]$}\;
  \For{$s \gets 1$ \KwTo $t$}{
    $\hat Q_{\text{new}}[s,t]\gets  - \frac{\alpha}{\chi^{tt}} \mathbb{E}\ab[G[s,t]]$
  }

\end{algorithm}

\section{Theoretical and Experimental Validation}\label{append:agreement}

\begin{figure*}[htb]
    \vskip 0.2in
    \begin{center}
    \centerline{\includegraphics[width=1\columnwidth]{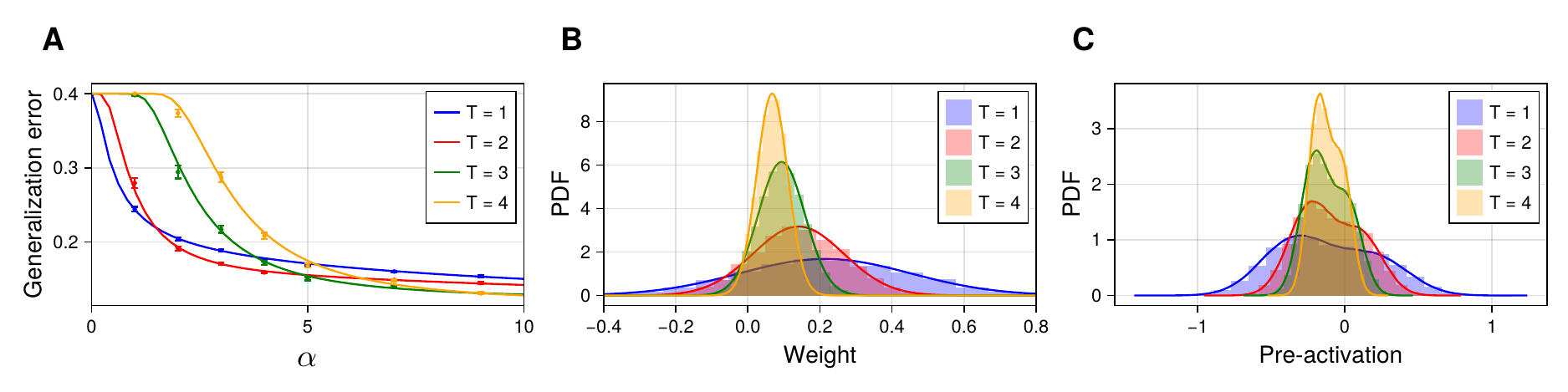}}
    \caption{Comparison of theoretical predictions derived from replica method and numerical experiments for the linear $t$-SD model statistics. 
    (A)
    Generalization error derived by the replica method (solid lines) and numerical simulations (dots with error bars).
    (B) 
    Distributions of optimal weights derived by the replica method (solid lines) and their empirical distributions obtained from a single experiment (histograms).
    (C) 
    Pre-activation distributions predicted by theory (solid lines) and empirically observed from a single experiment (histograms).
    Parameters for (A-C): $\rho = 0.4, \Delta = 0.6, \theta = 0.2, (\lambda_1, \lambda_2, \lambda_3, \lambda_4) = (1.5, 0.5, 2.0, 1.0), (\beta_1, \beta_2, \beta_3) = (0.8, 1.2, 1.0)$; (B, C) $\alpha = 3.0$.
    Numerical experiments: (A) $N = 10^3$ ( Error bars represent the standard error of the mean over 20 trials per point.); (B, C) $N = 10^4$.}
    \label{fig:theory_vs_experiment}
    \end{center}
    \vskip -0.2in
\end{figure*}

In this appendix, we present evidence demonstrating the strong agreement between theoretical predictions derived from the replica method and numerical experiments for the linear 
$t$-SD model. Figure~\ref{fig:theory_vs_experiment} compares the generalization error, weight distribution, and pre-activation distribution obtained from replica method with those from numerical experiments, revealing remarkable consistency between the two approaches.

\section{Optimization of the hyper parameters}
\label{appendix:hyperopt}
The above results describe the statistical properties of the estimators $\{\hat{\vb{w}}^t, \hat{B}^t\}_{t\ge0}$ and the generalization error for a fixed hyperparameters $\{\lambda^t\}_{t\ge0}, \{\beta^t\}_{t\ge1}$. In order to find the optimal hyper parameters, we used the Nelder-Mead (NM) method~\cite{vaibhav_kumar_dixit_2023_7738525}, which is a versatile black-box optimization algorithm. At each optimization stage in NM, we numerically solve the set of equations in Result~\ref{thm:full_theorem}, which can be efficiently solved using a simple fixed point iteration to evaluate the generalization error \eqref{eq:generalization_error}.

\section{Exact results for the linear $t$-SD model} 
\label{append:exact_solution}
In this appendix, we present several simplified analyses of the linear $t$-SD model and conclude by proving a generalized version of Result~\ref{prop:fixedpoint}.
\subsection{Integrated saddle point equations in the linear $t$-SD model}
In the case of the mean squared error loss and the linear activation function, the saddle point equations for the conjugate variables and $b^t$ is integrable.
For some algebraic manipulations, we have the following equations:
\begin{align}
  z^t_* &= \frac{\Delta\chi^{tt}}{2+\Delta\chi^{tt}}\ab(2y^{t-1}-h^t-1) \\
  y^t &= \frac{1}{2}\ab(\beta\ab(h^t+z^t_*) + 1) \quad (t \geq 1) 
\end{align}
\begin{align}
  \hat{Q}^{tt} &= \frac{\Delta\alpha}{2+\Delta\chi^{tt}} \\
  \hat{Q}^{st} &=  \frac{\Delta\alpha}{2+\Delta\chi^{tt}} \ab[-\beta^{t-1} \ab(\delta^{t-1,s}-\frac{1}{\alpha}\sum_{l=s}^{t-1}\chi^{l,t-1}\hat{Q}^{sl}) -  \frac{1}{\alpha}\sum_{l=s}^{t-1}\chi^{lt}\hat{Q}^{sl}] \quad (s < t) \\
\end{align}
\begin{align}
  \hat{m}^0 &= \frac{2\alpha\rho}{2+\Delta\chi^{00}}\ab(2(1-\theta)- (m^0+b^0)-1) \\
  \hat{m}^t &= \frac{2\alpha}{2+\Delta\chi^{tt}}\ab[\frac{\Delta}{2\alpha}\ab(\sum_{s=0}^{t-1}(\beta^{t-1} \chi^{s, t-1}-\chi^{st})\hat{m}^s )  + \beta^{t-1} \rho(m^{t-1}+b^{t-1})  - \rho(m^t+b^t)]  \quad (1 \leq t \leq T)
\end{align}
\begin{align}
  \hat{\chi}^{0t}
  &= \frac{\Delta}{2+\Delta\chi^{00}}
     \Bigl[
       \frac{2\alpha}{\Delta\chi^{tt}}\hat{r}^t
       + \Bigl(\sum_{l=0}^t Q^{0l}\hat{Q}^{lt} - m^0\hat{m}^t\Bigr)
     \Bigr], \\[8pt]
  \hat{\chi}^{st}
  &= \frac{\Delta}{2+\Delta\chi^{ss}}
     \Bigl[
       -\beta^{s-1}\Bigl(
         \sum_{l=0}^t Q^{\min(s-1,l),\max(s-1,l)}\hat{Q}^{lt}
         - \sum_{l=0}^{s-1}\chi^{l,s-1}\hat{\chi}^{lt}
         - m^{s-1}\hat{m}^t
       \Bigr) \\[4pt]
  &\qquad\quad
       + \Bigl(
         \sum_{l=0}^t Q^{sl}\hat{Q}^{lt}
         - \sum_{l=0}^{s-1}\chi^{ls}\hat{\chi}^{lt}
         - m^s\hat{m}^t
       \Bigr)
     \Bigr].
\end{align}
\begin{align}
  \hat{r}^{0} &= \frac{\Delta\chi^{00}}{2+\Delta\chi^{00}} \ab[(\rho+\theta-2\rho\theta) - \ab\{ \rho(1-\theta)(m^0+b^0) + (1-\rho)\theta(-m^0+b^0) \} ] \\
  \hat{r}^t &= \frac{\Delta\chi^{tt}}{2+\Delta\chi^{tt}}\left[ \sum_{s=0}^{t-1} \frac{1}{\chi^{ss}} \ab( \beta^{t-1}\chi^{s,t-1} - \chi^{st}) \hat{r}^s \right. \\
  & \qquad\qquad \left. + \beta^{t-1}\ab\{\rho(1-\theta)(m^{t-1}+b^{t-1})+(1-\rho)\theta(-m^{t-1}+b^{t-1})\} \right.  \\
  & \qquad\qquad \left. - \ab\{\rho(1-\theta)(m^{t}+b^{t})+(1-\rho)\theta(-m^{t}+b^{t})\} \right]   \quad (1 \leq t \leq T)
\end{align}
\begin{align}
  b^0 &= 2(\rho+\theta-2\rho\theta) - (2\rho-1)m^0 -1 \label{eq:b1} \\
  b^t &= \beta^{t-1} \ab((2\rho-1)m^{t-1} + b^{t-1}) - (2\rho-1)m^t  \quad (1 \leq t \leq T),  \label{eq:bt}
\end{align}
where we introduced the auxiliary variable $\hat{r}^t = \mathbb{E}_\mathcal{D}\ab[y^0 z^t_*]$ for simplicity.

\subsection{Case of $\rho = 1/2$ and $\lambda^0, \lambda^1\to\infty$}
One can solve the saddle point equations explicitly in some specific cases. 
For example, in the case of $\rho = 1/2$ and $\lambda^0, \lambda^1\to\infty$, the explicit form of the geneliztion error is given by 
following proposition.
\begin{proposition}
  In the linear $t$-SD model with $\rho = 1/2$ and $\lambda^0, \lambda^1 \to\infty$, the generalization errors at $t=0$ and $t=1$ are given by
  \begin{align}
    \mathcal{E}^0 &=  H\left(\frac{\sqrt{\alpha}(1-2\theta)}{\sqrt{\Delta\left(\Delta+\alpha(1-2\theta)^2\right)}}\right),\\
    \mathcal{E}^1  &= H\left( \frac{\alpha(\Delta+\alpha+\Delta\alpha)(1-2\theta)}{ \sqrt{ \Delta\left[ (\alpha^2+3\alpha+1)\Delta^3\alpha + \alpha^2\left(\Delta^2(\alpha^2+5\alpha+3) + \Delta(2\alpha+3)\alpha + \alpha^2\right)(1-2\theta)^2\right] }}\right).
  \end{align}
  In particular, $\mathcal{E}^{*0} = \mathcal{E}^0 $ and $\mathcal{E}^{*1} \leq \mathcal{E}^1$.
\end{proposition}

\subsection{Case of $\rho = 1/2$, $\lambda^0, \cdots, \lambda^T\to\infty$ and $T\to\infty$}
Another specific solvable case is $\lambda^0, \cdots, \lambda^T\to\infty$ and $T\to\infty$. 
For ease of reference, we first restate the generalized version of Result~\ref{prop:fixedpoint} as Result~\ref{prop:fixedpoint_general} below.  We then proceed to its proof.

\begin{result} (\textbf{The generalization error at $t\to\infty$})
    For an arbitrary choice of the set of the temperature parameters $\{\beta^t\}_{t\ge0}$,
    the generalization error of the linear $t$-SD model with $\rho=0.5$, $\lambda^0, \cdots, \lambda^t \to\infty$ and $t\to\infty$ is given by
    \begin{align}
        \lim_{t\to\infty} \mathcal{E}^t  &= \begin{cases}
        &0.5 \quad ( \alpha < \Delta^2) \\
        &H\ab(\sqrt{\frac{\alpha-\Delta^2}{\Delta(\alpha+\Delta)}}) \quad ( \alpha \geq \Delta^2) .
        \end{cases}
    \end{align}
    \label{prop:fixedpoint_general}
\end{result}

Under these conditions, equations \eqref{eq:b1} and \eqref{eq:bt} yield $b^0, \cdots, b^T = 0$.
For simplicity, we set $\lambda^0 = \cdots = \lambda^T = \lambda$, $\epsilon = 1/\lambda$, and $\gamma^1 = \cdots = \gamma^T = \gamma$ without loss of generality.
The scaling of each parameter with respect to $\epsilon$ and $\gamma$ is
\begin{align}
  m^t       &= \mathcal{O}\ab(\epsilon^t) \mathcal{O}\ab(\gamma^{t-1})     & \hat{m}^t &= \mathcal{O}\ab(\epsilon^{t-1}) \mathcal{O}\ab(\gamma^{t-1})\\
  Q^{st}    &= \mathcal{O}\ab(\epsilon^{s+t}) \mathcal{O}\ab(\gamma^{s+t-2}) & \hat{Q}^{st} &= \mathcal{O}\ab(\epsilon^{t-s-1})\mathcal{O}\ab(\gamma^{t-s}) \, (s < t), \, \hat{Q}^{tt} = \mathcal{O}\ab(1) \mathcal{O}\ab(1)\\
  \chi^{st} &= \mathcal{O}\ab(\epsilon^{t-s+1})\mathcal{O}\ab(\gamma^{t-s}) & \hat{\chi}^{st} &= \mathcal{O}\ab(\epsilon^{t+s-2}) \mathcal{O}\ab(\gamma^{t+s-2})\\
  R^{st}   &= \mathcal{O}\ab(\epsilon^{t+s-1}) \mathcal{O}\ab(\gamma^{t+s-2}).
\end{align}
Based on this scaling, we rescale each variable as 
\begin{align}
  m^t &\to \epsilon^t \gamma^{t-1} {m}^t  & \hat{m}^t &\to \epsilon^{t-1} \gamma^{t-1} \hat{m}^t \\
  Q^{st} &\to \epsilon^{s+t} \gamma^{s+t-2} Q^{st} & \hat{Q}^{st} &\to \epsilon^{t-s-1} \gamma^{t-s} \hat{Q}^{st} \, (s < t), \, \hat{Q}^{tt} \to \hat{Q}^{tt} \\
  \chi^{st} &\to \epsilon^{t-s+1} \gamma^{t-s} \chi^{st} & \hat{\chi}^{st} &\to \epsilon^{t+s-2} \gamma^{t+s-2} \hat{\chi}^{st} \\
  R^{st} &\to \epsilon^{t+s-1} \gamma^{t+s-2} R^{st}.
\end{align}
Taking the limit as $\epsilon \to 0$, we obtain the following simplified recurrence relations:
\begin{align}
  \chi^{st} &= \ab(\frac{\Delta\alpha}{2})^{t-s} 
\end{align}
\begin{align}
  \hat{Q}^{tt} &= \frac{\Delta\alpha}{2}, \quad  \hat{Q}^{t-1,t} = -\frac{\Delta\alpha}{2} \\
  \hat{Q}^{st} &= -\ab(\frac{\Delta(1+\alpha)}{2})^{t-s-2} \frac{\Delta^2\alpha}{4} \quad (t\geq s\geq 2)
\end{align}
\begin{align}  
  \hat{m}^0 &= m^0 =  \frac{\alpha}{2}(1-2\theta)  \\
  \hat{m}^1 &= \frac{1}{2}(\Delta+\alpha)m^0 , \quad m^1 = \frac{1}{2}(\Delta+\alpha + \Delta\alpha)m^0 \\
  \hat{m}^t &= \frac{\Delta}{2}\ab(\frac{\Delta\alpha}{2})^{t-1}\sum_{s=0}^{t-1}\ab(\frac{\Delta\alpha}{2})^{-s}\hat{m}^s + \frac{\alpha}{2}m^{t-1} \quad (t\geq 1) \label{eq:mt} \\
  m^t &= \frac{\Delta^2\alpha}{4} \ab(\frac{\Delta}{2}(1+\alpha))^{t-2} \sum_{s=0}^{t-2} \ab(\frac{\Delta}{2}(1+\alpha))^{-s}m^s + \hat{m}^{t} + \frac{\Delta\alpha}{2}m^{t-1} \quad (t\geq 2) \label{eq:mt2}
\end{align}
\begin{align}
  R^{s0} &= \hat{\chi}^{0s} \\
  R^{st} &= \hat{\chi}^{\min\{s,t\},\max\{s,t\}} - \sum_{l=0}^{t-1}\hat{Q}^{lt}R^{sl} \quad (t\geq 1) \label{eq:Rst}
\end{align}
\begin{align}
  Q^{0t} &= \hat{m}^0m^t + R^{0t} \\
  Q^{st} &= \hat{m}^sm^t + R^{st}  + \frac{\Delta^2\alpha}{4}\ab(\frac{\Delta}{2}(1+\alpha))^{s-2} \sum_{l=0}^{s-2} \ab(\frac{\Delta}{2}(1+\alpha))^{-l}Q^{lt} + \frac{\Delta\alpha}{2}Q^{s-1,t} \quad (t\geq s\geq 1) \label{eq:Qst}
\end{align}
\begin{align}
  \hat{\chi}^{00} &= \frac{\Delta\alpha}{4} \\
  \hat{\chi}^{0t} &= \frac{\Delta}{2}\ab(\sum_{s=0}^{t-1} \chi^{s,t-1}\hat{\chi}^{0s} + \frac{\alpha}{2}(1-2\theta)m^{t-1}) \quad (t\geq 1) \\
  \hat{\chi}^{st} &= -\frac{\Delta}{2} \ab(\sum_{l=0}^{t-1} Q^{\min\{s-1,l\}, \max\{s-1,l\}}\hat{Q}^{lt} - \sum_{l=0}^{s-1}\chi^{l,s-1}\hat{\chi}^{lt} - m^{s-1}\hat{m}^t )  \quad (t\geq s\geq 1) .\label{eq:chist}
\end{align}

We now consider the solution to these recurrence relations for sufficiently large $t$ ($t \gg 1$). Let us propose a trial solution of the form $m^t = c \mathcal{M}^t, \hat{m}^t = c L \mathcal{M}^t$, where $\mathcal{M}>\Delta(1+\alpha)/2$ and $c$ is a constant depending on the initial condition $\theta$. 
Note that $t$ in the left-hand side denotes the step number, while $t$ in the right-hand side is an exponent. 
We have:
\begin{align}
  \sum_{s=0}^{t-1} \ab(\frac{\Delta\alpha}{2})^{-s} \hat{m}^s  &=  c \sum_{s=0}^{t-1} \ab(\frac{\Delta\alpha}{2})^{-s} L \mathcal{M}^s  + \mathcal{O}(1)\\
  \sum_{s=0}^{t-2} \ab(\frac{\Delta}{2}(1+\alpha))^{-s} m^s &= c \sum_{s=0}^{t-2} \ab(\frac{\Delta}{2}(1+\alpha))^{-s} \mathcal{M}^s + \mathcal{O}(1)
\end{align}

From equations \eqref{eq:mt} and \eqref{eq:mt2}, we obtain the solution satisfying the condition $\mathcal{M}>\Delta(1+\alpha)/2$:

\begin{align}
  m^t &= c (1+\Delta) \mathcal{M}^t \\
  \hat{m}^t &= c \mathcal{M}^t
\end{align}    
where 
\begin{align}
  \mathcal{M} &= \frac{1}{4} \ab(2\alpha\Delta+\alpha+\Delta + \sqrt{\alpha^2 + 2\alpha\Delta(2\Delta+1)+\Delta^2})  .
\end{align}

Next, we consider solutions of the form $Q^{st} = c^2 q \mathcal{M}^{s+t}, R^{st} = c^2 r \mathcal{M}^{s+t}, \hat{\chi}^{st} = c^2 \chi \mathcal{M}^{s+t}$ for $s,t \gg 1$. Substituting these into \eqref{eq:Rst}, we obtain
\begin{align}
  r &= (1+\Delta) \chi \\
  q &= (1+\Delta)^2 (1+\chi) \\
  \chi &= \frac{\Delta}{\alpha}\ab(1 + \frac{\Delta}{\ab(1+\Delta)^2}q) .
\end{align}
Solving these equations yields
\begin{align}   
  r = \frac{\Delta(1+\Delta)^2}{\alpha-\Delta^2}, \quad q = \frac{(1+\Delta)^2(\alpha+\Delta)}{\alpha-\Delta^2}, \quad \chi = \frac{\Delta(1+\Delta)}{\alpha-\Delta^2} .
\end{align}
Consequently, the generalization error as $t\to\infty$ is given by
\begin{align}
  H\ab(\frac{m^t}{\sqrt{\Delta Q^{tt}}}) \to  H\ab(\sqrt{\frac{\alpha-\Delta^2}{\Delta(\alpha+\Delta)}}) .
\end{align}

However, when $\alpha < \Delta^2$, this solution becomes inappropriate as $Q^{tt} < 0$. In this case, the scale $\mathcal{N}$ of $Q^{st} = c^2q\mathcal{N}^{s+t}$ satisfies $\mathcal{N}>\mathcal{M}$, and the generalization error becomes  $H(0)=0.5$. This completes the proof of Result~\ref{prop:fixedpoint_general}.

\begin{figure}[htb]
    \centering
    \includegraphics[width=0.6\linewidth]{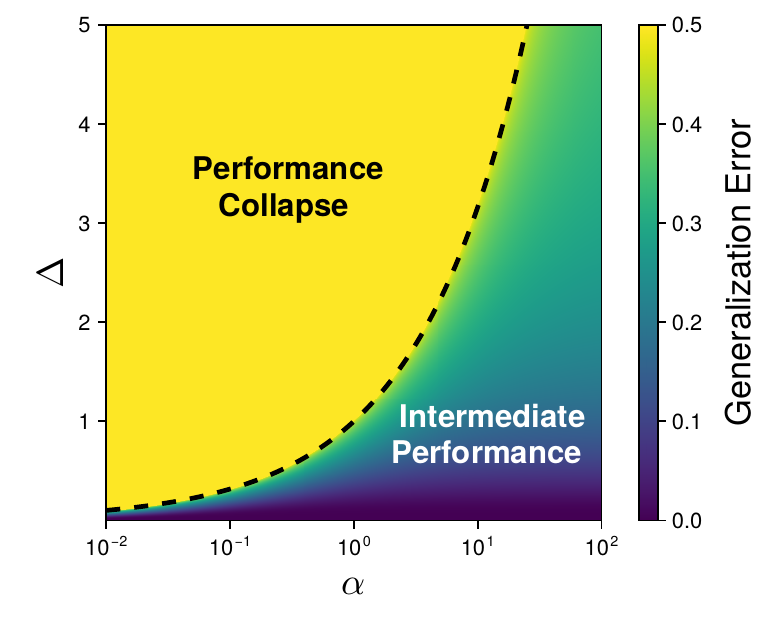}
    \caption{Theoretical prediction of generalization error for the linear $t$-SD model with $\lambda^0, \cdots, \lambda^t \to\infty$ and $t\to\infty$ with
    the phase transition boundary indicated by the dashed line.}
    \label{fig:phase_tran}
\end{figure}

Result~\ref{prop:fixedpoint_general} reveals a phase transition phenomena at $\alpha=\Delta^2$. A generalization error of 0.5 means that the performance of $t$-SD
is equivalent to random guessing; hence we refer to the phase $\alpha<\Delta^2$ as the performance collapse phase. 
The independence from the choice of the temperature is natural since it only affects on the scale of the weight in the linear $t$-SD model.
In Figure~\ref{fig:phase_tran}, dependence of ${\cal E}^t$ on $\alpha$ and $\Delta$ at $t\to\infty$ is shown, with phase transition boundary represented by dashed line.
The generalization error at $t \to \infty$ for $\alpha \geq \Delta^2$ is below 0.5, indicating performance better than random guessing, but it remains higher than the optimal error; hence, we refer to the phase $\alpha \geq \Delta^2$ as the intermediate performance phase.

\section{Experimental Details}\label{append:experiment}
This appendix provides the detailed experimental settings employed in Section~\ref{sec:experiments}.
All experiments were executed on CPU workers equipped with an AMD EPYC 9654 processor and 512 GB of main memory. Each run required approximately two hours of wall-clock time using all available cores.

\subsection{Data and backbone selection}
We consider the binary `cat vs. dog' subset of CIFAR-10~\cite{Krizhevsky2009-ac} (licensed under the MIT License) and employ two deep backbones, ResNet-18 and ResNet-50, pre-trained on ImageNet.

\subsection{Feature Extraction}
Each CIFAR-10 image is first resized and normalized to match the preprocessing used during ImageNet training.  We then remove the backbone's final classification layer and take the output of the penultimate layer as a fixed embedding of dimension $N=512$ for ResNet-18 or $N=2048$ for ResNet-50~\cite{torchvision2016} (licensed under BSD 3-Clause ``New'' License).  
All embeddings and their clean labels are saved for downstream use.

\subsection{Label noise injection}
To simulate noisy supervision, we flip each training label independently with probability $\theta$.  Test labels remain untouched.

\subsection{Training subset sampling}
From the pool of noisy embeddings, we uniformly draw $M$ samples (with a fixed class balance when desired) to form the actual training set used in the SD experiments.

\subsection{Self-distillation and hyperparameter tuning.}
Using those $M$ examples, we perform the logistic $t$-SD procedure defined in Section~\ref{sec:problem_setup}.  
The key hyperparameters, $\lambda$ and $\beta$ are selected by minimizing the estimated generalization error on the test embeddings via Bayesian optimization.  
Concretely, we model the test error as a function of ($\lambda$,\ $\beta$) with a Gaussian‐process surrogate and optimize its expected improvement.

\section{Additional experiments}\label{append:additional_exp}

\subsection{Self-distillation on a noiseless dataset}
\begin{figure}[htb]
    \centering
    \includegraphics[width=0.9\linewidth]{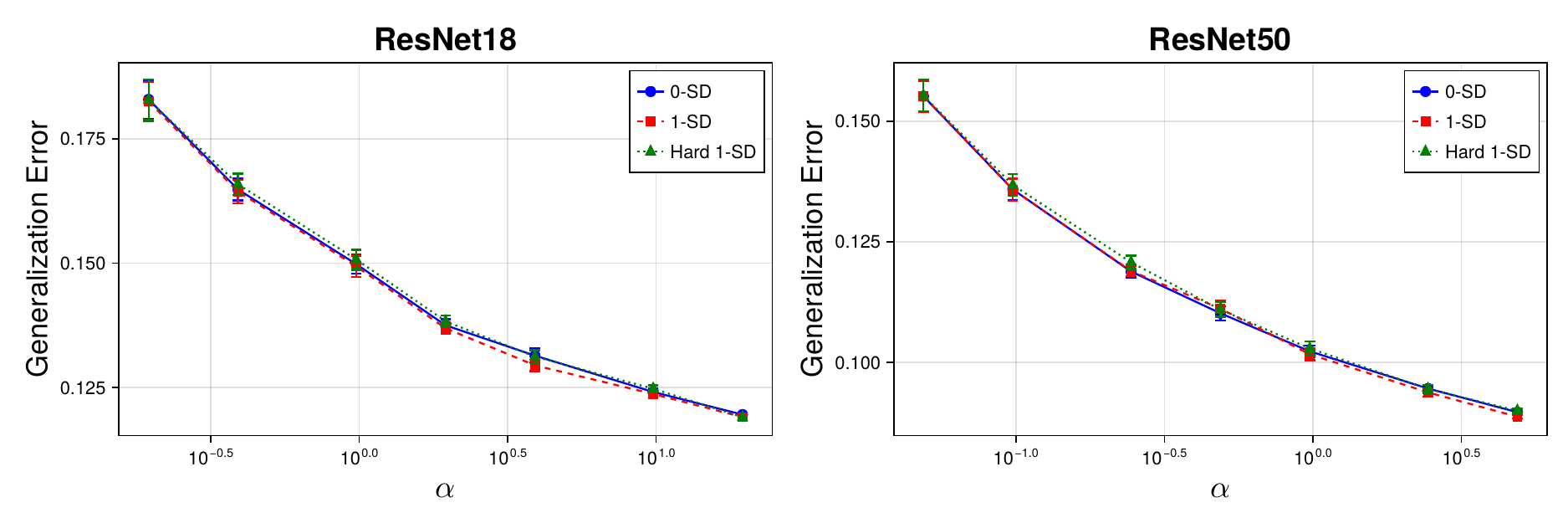}
    \caption{Comparison of the optimal generalization error of the logistic $0$-SD model, $1$-SD model and $1$-SD model using hard pseudo labels for CIFAR-10 dog versus cat classification using pretrained ResNet-18 ($N=512$) and ResNet-50 ($N=2048$) feature representations. Parameters: $\theta=0.0$. Error bars represent the standard error of the mean over $10$ trials per point.}
    \label{fig:noiseless_exp}
\end{figure}

We repeat our $1$-SD experiments with no label noise ($\theta=0$) to isolate the denoising effect from the dark knowledge effect.
Since there is no label noise, any gain in generalization must arise solely from the teacher's soft outputs.
As shown in Figure~\ref{fig:noiseless_exp}, no meaningful improvement is observed under realistic settings.

These findings validate the hypothesis of Section~\ref{sec:soft_labels}: in linear models under the Gaussian mixture model, dark knowledge alone yields only marginal benefit. The dominant mechanism by which SD enhances performance may be denoising, not the transmission of refined probability information even in more realistic scenarios.

\subsection{Optimal soft labels for self-distillation}
\begin{figure}[htb]
    \centering
    \includegraphics[width=1.0\linewidth]{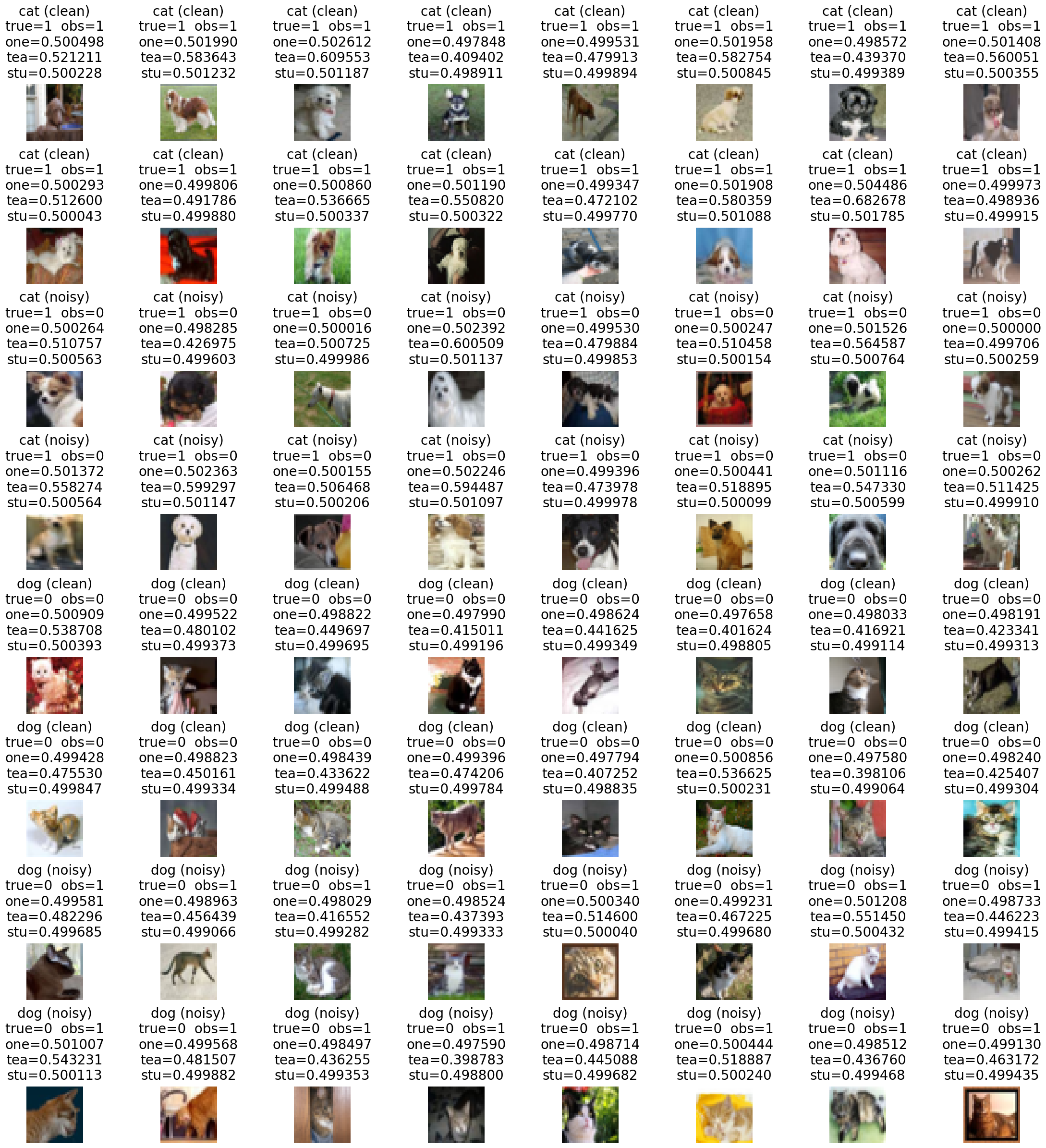}
    \caption{Sample images from the CIFAR-10 training set used in our experiments. `true' and `obs' indicate the ground-truth and noisy observed labels (0 for dog, 1 for cat), respectively. `one', `tea', and `stu' denote the predicted labels under the optimal $0$-SD model, the pseudo-labels provided to the student by the optimal $1$-SD teacher, and the predicted labels under the optimal $1$-SD student, respectively. Parameters: $\theta=0.4, M=1000, N=512$; ResNet-18 features and logistic $t$-SD.}
    \label{fig:cifar_picture}
\end{figure}

A natural question is whether the soft labels produced by the optimal $1$-SD teacher simply mirror the probabilities predicted by the optimal $0$-SD model, or whether they differ in a systematic way. Figure~\ref{fig:cifar_picture} shows, for several randomly chosen training samples, the ground-truth label, the noisy observed label, the predicted probability under optimal $0$-SD, the pseudo-label assigned by the optimal $1$-SD teacher, and the student's prediction under optimal $1$-SD. The optimal teacher consistently issues more extreme confidence scores than the base model.
This observation suggests that the most effective labels for student improvement can deviate from the model's optimal activations.

\section{Further remarks on limitations and future works}\label{append:further_remarks}
In this appendix, we outline several promising avenues that address the limitations of our current study and extend its insights.

Our theoretical analysis assumes purely linear models under Gaussian‐mixture noise and relies on asymptotic $N,M\to\infty$ formulas, so its accuracy may degrade on finite‐sample, non‐linear settings.  Likewise, although our CIFAR-10 probes demonstrate feasibility, they do not guarantee performance on larger or more complex vision tasks.  Addressing these gaps suggests following several natural directions for further work.

\subsection{Extension to anisotropic data distributions} 
In anisotropic settings, the impact of SD is inherently direction-dependent: it tends to amplify useful signals along high-variance directions, while potentially neglecting or even distorting low-variance ones. 
As a consequence, the overall benefit of SD may become larger when the task-relevant information aligns with principal components, but smaller when crucial information lies in weak directions. 
Understanding this interplay between pseudo-label dynamics and the spectral structure of the data covariance is an important direction for future work.

\subsection{Extension to other distillation strategies} 
    While this study focuses on linear models, extending the analysis to deep learning presents promising directions for future research. In deep models, dark knowledge may differ significantly and hold greater significance than in linear analysis, due to their feature learning capabilities. 
    For instance, models propagating intermediate layer information~\cite{Zhang2019-lq} might depend more on transfer of feature representations rather than predictions alone.
    Additionally, another avenue lies in exploring the interplay between SD and security, particulaly optimizing defense against model stealing attacks~\cite{Ma2021-ga, 11014106}.
    This line of inquiry extends our problem setting to a min-max framework, aiming to minimize the effectiveness of SD. Advancing these directions could contribute to both KD robustness and secure machine learning. 
    
\subsection{What are the best pseudo-labels to learn?} 
    Because the optimization goal of $t$\text{-SD} is the generalization error of the final ($t$-th) student, every intermediary teacher must issue labels not to mirror the true decision boundary but to maximize the downstream student's performance.
    Intriguingly, these student-centric labels differ substantially from the teacher-centric labels that would be chosen for standard classification, and more confident soft labels tend to produce stronger students. 
    A more detailed analysis of these label distinctions is likely to reveal the answer to the fundamental question of which labels are most beneficial for learning. 
    
\subsection{Application of multi-stage replica theory to other learning problems} 
    We believe that the multi-stage replica method we employed for theoretical analysis can be extremely useful for probing learning dynamics that have, to date, remained inaccessible to DMFT.  
    Problem formulations in which the outcomes of one learning phase are reused in a subsequent phase setting ups to which our framework can be applied directly include domain adaptation~\cite{Blitzer2006-vu}, curriculum learning~\cite{Bengio2009-gy} and meta-learning~\cite{Hospedales2022-ap}.  
    It would also be intriguing to investigate the generalization-error dynamics of deep neural networks under a regime where layers are trained sequentially and iteratively~\cite{pmlr-v235-cui24d}.

\end{document}